\documentclass[12pt,MSc]{muthesiss}
\usepackage[utf8]{inputenc}

\usepackage{amssymb}
\usepackage{graphicx}
\usepackage{amsmath}
\usepackage{algorithm}
\usepackage{algpseudocode}
\usepackage{epigraph}
\usepackage{amsmath,mleftright}
\usepackage{xparse}
\usepackage{multirow}
\usepackage{booktabs}
\usepackage{listings}
\usepackage{setspace} 
\usepackage{color}
\usepackage{hyperref}
\usepackage[english]{babel}
\newtheorem{theorem}{Theorem}

\begin{document}
\title{An Analysis of \\Physics-Informed \\Neural Networks}
\author{Edward Small}
% Faculty of Life Sciences people should comment the next line out
\school{Mathematics}
\faculty{Science and Engineering}
\def\wordcount{13637}

% Uncomment the line below to suppress the `List of Tables' page (optional)
%\tablespagefalse

% Uncomment the line below to suppress the `List of Figures' page (optional)
%\figurespagefalse

% Uncomment the line below to use a customised Declaration statement
%\def\declaration{All the work in this thesis has been sourced from Google.}
\beforeabstract

Whilst the partial differential equations that govern the dynamics of our world have been studied in great depth for centuries, solving them for complex, high-dimensional conditions and domains still presents an incredibly large mathematical and computational challenge. Analytical methods can be cumbersome to utilise, and numerical methods can lead to errors and inaccuracies. On top of this, sometimes we lack the information or knowledge to pose the problem well enough to apply these kinds of methods.

Here, we present a new approach to approximating the solution to physical systems - physics-informed neural networks. The concept of artificial neural networks is introduced, the objective function is defined, and optimisation strategies are discussed. The partial differential equation is then included as a constraint in the loss function for the optimisation problem, giving the network access to knowledge of the dynamics of the physical system it is modelling.

Some intuitive examples are displayed, and more complex applications are  considered to showcase the power of physics informed neural networks, such as in seismic imaging. Solution error is analysed, and suggestions are made to improve convergence and/or solution precision. Problems and limitations are also touched upon in the conclusions, as well as some thoughts as to where physics informed neural networks are most useful, and where they could go next.

\afterabstract

% The next part is optional; however it is a good place to thank your
% supervisor and the people responsible for providing computer support ;-)

\addcontentsline{toc}{chapter}{Notations}
\chapter*{Notations}
\begin{tabular}{p{4cm}p{11cm}}
$\alpha, \beta, \gamma,...$ & Lowercase Greek letters used for scalars $\mathbb{R}$ \\
$a, b, x, y, z$ & Used for vectors $\mathbb{R}^n$ \\
$x_i$ & A generic element inside the vector $x$ \\
$x_k$ & State of vector $x$ in iteration $k$ of an algorithm \\
$\hat{y}$ & An estimate of the true output $y$ \\
$A, B, C, W, M$ & Used for matrices $\mathbb{R}^{n \times m}$ or sets \\
$A_{i, j}$ & A submatrix inside the matrix $A$ \\
$f, g, \phi, \sigma$ & Functions mapping one set to another \\
$f_t, \phi_{yy}, u_x$ & Functions that have been differeniated w.r.t. the subscript \\
$f_n, \phi_n$ & The nth function in a series or set \\
$F, G$ & ANN approximation the lowercase functions $f, g$ \\
$L$ & Number of layers in a network \\
$l$ & A specific layer in a network \\
$N$ & The number of elements in a set or series \\
$n^{[l]}$ & A specific neuron in a specific layer \\
$\mathcal{L}$ & Loss function \\
$|| x ||_p$ & vector norm $\big ( \sum_i|x_i|^p\big)^\frac{1}{p}$ \\
$\mathcal{J}$ & The Jacobian \\
$\mathcal{H}$ & The Hessian \\
$\mathcal{N}$ & A differential operator \\
$\Lambda$ & A physics-informed neural network \\
$\theta$ & Values that parameterise a ANN, such as weights and biases
\end{tabular}

\clearpage
\newpage

\addcontentsline{toc}{chapter}{Abbreviations}
\chapter*{Abbreviations}
\begin{tabular}{p{4cm}p{11cm}}
AD & Automatic Differentiation \\
ADAM & Adaptive Moment Estimation \\
ANN & Artificial Neural Network \\
AutoGrad & Adaptive Stochastic Gradient Descent \\
BC & Boundary Condition \\
CPU & Central Processing Unit \\
FD & Finite Difference \\
FEA & Finite Element Analysis \\
FWI & Full Waveform Inversion \\
GPU & Graphics Processing Unit \\
IBM & International Business Machine \\
IC & Initial Condition \\
KdV & Korteweg–De Vries \\
ML & Machine Learning \\
MRI & Magnetic Resonance Imaging \\
MSE & Mean Squared Error \\
NLP & Natural Language Processing \\
L-BFGS & Limited memory Broyden–Fletcher–Goldfarb–Shanno \\
LU & Lower Upper Factorisation \\
ODE & Ordinary Differential Equation \\
PDE & Partial Differential Equation \\
PINN & Physics-informed Neural Network \\
RMSProp & Root Mean Square Propogation \\
SAS & Statistical Analysis Software \\
SGD & Stochastic Gradient Descent \\
s.t. & Such that \\
TV & Total Variation \\
WRI & Wavefield Reconstruction Inversion \\
w.r.t & With respect to
\end{tabular}

\prefacesection{Acknowledgements}
I would like to express my immense gratitude to my supervisor Dr Oliver Dorn for the vast amount of experience and wisdom he supplied whilst writing this thesis. He was a fantastic source of inspiration for much of my time at the University of Manchester, and his guidance was crucial for the research I have conducted. My enjoyment of his lecture course in numerical optimisation and inverse problems was what originally motivated me to study this topic.

Of course, I also have to think about those who were immediately around me. Whether it was my mother Annette, who would bring me the occasional coffee when I was deep in thought, my father Graham, who always had time to listen to me, or my friend Matt, who was a good reminder that taking a break can be a good thing, I will be forever grateful. I especially appreciated them nodding along as I excitedly explained what I was researching, despite none of them really understanding what I was saying.

And finally, I have to thank one particularly special person who couldn't be around me, but still offered me constant, endless support - my partner Jess. Australia has never felt so far away as it has for the last 2 years, but here is hoping we will see each other soon.

% The next line is NOT optional and MUST appear
\afterpreface
\pagenumbering{arabic}

% Finally, you can start writing about all the new theorems you have proved
% and all the new results that you have discovered

\chapter{Introduction}

Machine Learning is a relatively young and fast developing branch of mathematics. Whilst the first mention of machine learning wasn't until 1952 by IBM computer scientist Arthur Samuel (who was developing complex algorithms for computers to effectively play checkers \cite{checkers}), the foundations of machine learning are scattered throughout modern history, with its inception standing on the shoulders of many well-known mathematicians, such as Laplace, Markov \cite{markov1, markov2}, Bayes \cite{bayes}, and Turing \cite{turing}, to name a few. It poses the elegant question \textit{`what if my computer could learn?'} That is to say, how could one create an algorithm that was dynamic in its problem solving approach - an algorithm that could learn from its mistakes, and iteratively adapt and evolve to become more accurate?

Modern mathematicians have spent decades refining these learning processes and, coupled with the age of \textit{Big Data}\footnotemark, these algorithms have been put to incredible use. From facial recognition software \cite{face}, to self-driving cars \cite{drive} and targeted advertisement \cite{advert}, machine learning algorithms have consumed vast amounts of data to create discrete solutions to problems that no human ever could.

One recent advancement in a sub-branch of machine learning, artificial neural networks (ANNs), is to use the known the physics of a system as a constraint for optimising the learning process, ultimately finding approximate solutions to partial differential equations (PDEs). Whilst this process is somewhat complex, it can not only lead to more accurate results, but also achieve these results at a higher computational speed. These special types of network constraints are known as \textit{physics-informed neural networks} (PINNs). 

\footnotetext{SAS, one of the largest data management and analytics companies in the world, defines \textit{Big Data} as datasets that are so large, fast (large amounts of new data generated per time step) or complex (highly variable with high veracity) that analysing the data cannot be done through traditional statistical methods. This kind of data poses a unique challenge in terms of processing and storage.}

\newpage
\section{What is an Artificial Neural Network?}

PINNs are a subtype of ANNs, and so before learning about PINNs it is important to have a basic understanding of ANNs - how they process data, produce results, and how they learn.

\subsection{The Network}

\begin{figure}[b]
\begin{center}
\includegraphics[width=0.45\textwidth]{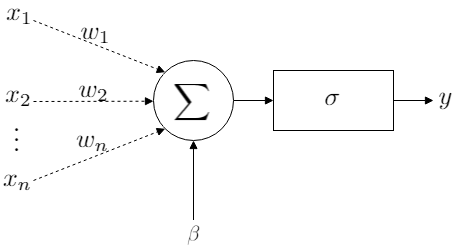}
\caption{Visual representation of a single perceptron taking in $n$ inputs.}
\label{di:perceptron}
\end{center}
\end{figure}

As the name suggests, ANNs are heavily inspired by the basic operations and understanding of a neuron - the nerve cell that is the building block of the nervous system. A single artificial neuron is known as a \textit{perceptron}\cite{perceptron}. A perceptron takes a set of inputs $x \in \mathbb{R}^n$ and gives a single output $y \in [-1, 1]$, where $y$ is a measure of `how active' the perceptron is. A perceptron achieves this by applying 3 fundamental steps to the inputs:

\begin{enumerate}
    \item Sum the product of each input $x_i \in x$ with a corresponding weight $w_i \in w$. Each weight $w_i$ can be interpreted as a measure of how sensitive a perceptron is to each individual input $x_i$. As an example, if $w_i = 0$ then the input $x_i$ has no influence on the activation of the perceptron.
    \item Add a bias $\beta$ to this sum. This bias can be interpreted as a measure of how active a neuron would be if $x_i=0$ for $i=1,...,n$.
    \item Put this value through an activation function $\sigma: \mathbb{R} \to [-1, 1]$, such as $\tanh$.
\end{enumerate}
The activation of a perceptron $a$ can therefore be represented as
\begin{equation}
    a = \sigma \bigg( \sum_{i=1}^n (x_i w_i) + \beta \bigg)
    \label{eq:perceptron}
\end{equation}
which visually looks like Figure \ref{di:perceptron}.

 Intuitively then, an ANN $F$ (which approximates $f$) is just a collection of these perceptrons feeding into each other \cite{multi-perc}, the activation of previous neurons influencing the activation of others such that $F: x \to \hat{y}$. Typically, an ANN is organised into $l=1, 2, ..., L$ layers, where each layer has $n^{[l]}$ number of neurons. Any layer between the input layer $n^{[1]}$ and the output layer $n^{[L]}$ is usually referred to as a \textit{hidden layer} because its state is not accessible by a user \cite{hiddenlayer}. The user only has access to the input layer and output layer.

\begin{figure}[t]
\begin{center}
\includegraphics[width=0.55\textwidth]{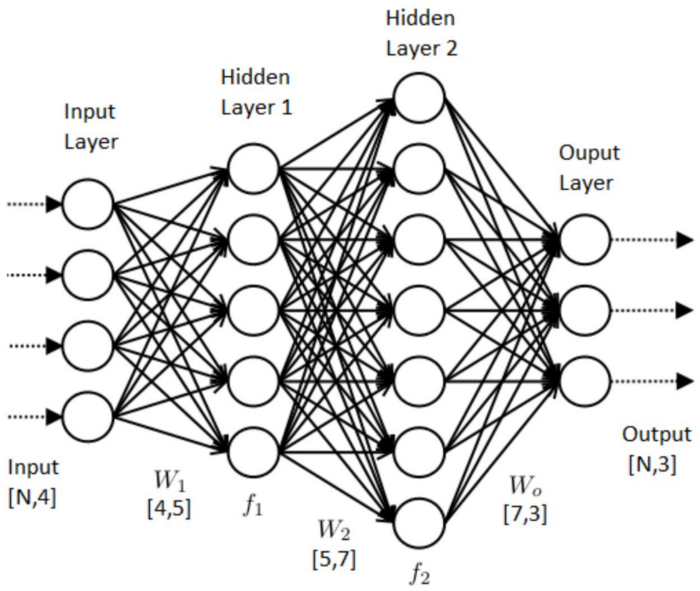}
\caption{An ANN with $x \in \mathbb{R}^4$ inputs, 2 hidden layers, and an output layer $\hat{y} \in \mathbb{R}^3$}
\label{di:ann}
\end{center}
\end{figure}

For $l=2, ..., L-1$ the activation $a^{[l]}$ is therefore

\begin{equation}
    \begin{bmatrix}
        a^{[l]}_1 \\
        a^{[l]}_2 \\
        \vdots \\
        a^{[l]}_{n^{[l]}}
\end{bmatrix}
 =
\sigma
\bigg(
    \begin{bmatrix}
        w_{1,1} & w_{1,2} & \dots & w_{1, n^[l-1]}\\
        w_{2,1} & w_{2,2} & \dots & w_{2, n^[l-1]} \\
        \vdots & \vdots & \ddots & \vdots \\
        w_{n^{[l]}, 1} & w_{n^{[l]}, 2} & \dots & w_{n^{[l]}, n^{[l-1]}}
\end{bmatrix}
    \begin{bmatrix}
        a^{[l-1]}_1 \\
        a^{[l-1]}_2 \\
        \vdots \\
        a^{[l-1]}_{n^{[l-1]}}
\end{bmatrix}
+
\begin{bmatrix}
        b^{[l]}_1 \\
        b^{[l]}_2 \\
        \vdots \\
        b^{[l]}_{n^{[l]}}
\end{bmatrix}
\bigg) \\
\end{equation}
which can be condensed into
\begin{equation}
    a^{[l]}=\sigma(W^{[l]}a^{[l-1]} + b^{[l]})
\end{equation}
where $a^{[l]},  b^{[l]} \in \mathbb{R}^{n^{[l]}}$ and $W^{[l]} \in \mathbb{R}^{n^{[l]} \times n^{[l-1]}}$. This is the key to how ANNs propagate information through the system to find a mapping for $f:x \to y$. 

Neurons can be connected in many different ways to create special types of networks, such as \textit{convolutional networks}\footnote{A convolutional network has a sense of locality or grouping in the input data in such a way that not every neuron in a layer connects to every neuron in the next.} or \textit{recurrent networks}\footnote{A recurrent network retains some memory of previous inputs and outputs in such a way that previous entries can influence future results.}. Figure \ref{di:ann} is an example of a \textit{fully connected} ANN, as every node in layer $l-1$ connects to every node in $l$. For simplicity, we will mostly consider fully connected, feedforward ANNs, unless otherwise specified.

An important result from ANNs is that a network $F$ with a single hidden layer can approximate any continuous function $f$ within a region $I_n$ (which is n-dimensional) to an arbitrary precision $\epsilon > 0$ such that
\begin{equation}
    |F(x) - f(x)| < \epsilon \qquad \forall \quad x \in I_n
\end{equation}
provided that no limit is placed on the values of $W$ and $b$ (the weights and biases of the ANN $F$). The precision, therefore, is dependent on the amount of neurons. This is known as the \textit{universal approximation theorem}\cite{universal}.

\section{What is a Partial Differential Equation?}

A partial differential equation (PDE) is a general equation that describes the relationship between rates of changes between variables in a multi-variable function \cite{pde}. A famous example is the 1-dimensional heat equation, which describes the heat flux on a rod with 1 dimension, $x$, through time, $t$, such that 
\begin{equation}
    u_t = k u_{xx}
\end{equation}
where $u_t = \frac{\partial u}{\partial t}$ is the change in temperature through time, $u_{xx}=\frac{\partial^2 u}{\partial x^2}$ is the spatial change in the temperature through space, and $k$ is a convective term that describes how easily energy can flow in/out of the rod \cite{heat}. The equation relates the change in temperature over time at a specific point in space and time to the change in temperature through space at a specific point in space and time.

PDEs are used generally to describe all kinds of physcial systems, such as acoustics, diffusion, electromagnatism, fluid dynamics, and even quantum mechanics \cite{pdeuse}, so finding solutions to these equations (given certain initial conditions (ICs)\footnote{If we know the initial conditions then we know the general shape of the function for all variables at time $t=0$} and boundary conditions (BCs)\footnote{If we know the boundary conditions, we know what the behaviour of the function should be for all time $t$ at the extreme parts of the domain of interest.}) is an incredibly important mathematical endeavour. Unfortunately, finding solutions to PDEs of real physical phenomena  can be incredibly challenging for a multitude of reasons. 

There are two main types of solutions: \textit{analytical solutions} and \textit{numerical solutions}. Analytical solutions can be thought of as closed-form solutions that match both the PDE, and the ICs and BCs of a problem, sometimes uniquely (depending on how the problem is defined). Solving a PDE analytically can be an incredibly time-intensive task, involving classifying the PDE and then utilising intricate methods, such as periodic extensions, Fourier series \cite{fourier}, and scattering \cite{scattering}. Even then, there is no guarantee that the solution will be easy to use, and sometimes the general solution is in the form of a complex integral, or an infinite series
\begin{equation}
    \phi(x, t) = \sum_{n=1}^{\infty} \alpha_n \phi_n(x, t) 
\end{equation}
where each $\phi_n$ is a solution to the PDE. Thus, even if $\alpha_n \to 0$ as $n \to \infty$ rapidly, we must accept that we can only estimate the true solution to an arbitrary precision. 

Finding analytical solutions in real-world modelling is incredibly challenging for other reasons too. These challenges include not knowing the exact ICs and BCs of a problem, or having a data-set that is too sparse, or a domain that is too complex to accurately define in a usable way. The problem deepens considerably when one considers the fact that some widely used PDEs, such as the famous Navier-Stokes (NS) equations describing the motion of fluid, are actually not well understood. In fact, we do not even know if smooth solutions always exist for 3 dimensional fluid flow, and turbulence and singularities still pose a huge problem \cite{ns}. Since the NS equations govern the dynamics of physical phenomena like atmospheric dynamics, and therefore weather predictions, it should be clear why study of these equations is paramount.

Numerical methods can help with this, such as finite difference (FD) and finite element analysis (FEA). FD involves estimating the value of derivatives to estimate the value of the function at the next point in space and time, whereas FEA uses variational methods to produce an approximate solution to the boundary conditions across individual, easier to solve elements, and then stitching the solution together using interpolation.

These techniques can have their own issues. For example, the errors in FD methods are propagated into the next iteration, and so using them for large domains for large time is somewhat challenging \cite{fderror}. FEA can have continuity issues when crossing over elements, called flux jumps, that lead to energy loss \cite{feerror}. Accurate solutions for either method in complex domains often relies on lot of computing power, and if we want to know the value of a function at a particular point in space and time, we often need to know the value of the surrounding points in space and time - something that is not necessary for an analytical solution.

\chapter{Training a Network to Approximate a Function}

The goal of an ANN may be to replicate the behaviour of an unknown function $f:\mathbb{R}^n \to \mathbb{R}^m$. In machine learning there are three main types of learning processes
\begin{enumerate}
    \item Supervised learning.
    \item Unsupervised learning\footnote{Unsupervised learning is used when the inputs for a function are known, but the corresponding outputs are not. The goal, therefore, is usually to find classes, groups or structures in the input data. This type of learning is often used in \textit{dimension reduction}}.
    \item Reinforcement learning\footnote{Reinforcement learning is used when a network learns through a penalty/reward system. Given a certain goal the network should optimise itself to produce the largest reward with the smallest penalty}.
\end{enumerate}
Supervised learning is used when a labelled data set is available \cite{suplearning}. That is to say, whilst the true function $f$ may be unknown, we do have access to a set of inputs $X \in \mathbb{R}^{n \times p}$ and matched outputs $Y \in \mathbb{R}^{m \times p}$ where, if $x \in X$ is a column in $X$ and $y \in Y$ is a corresponding column in $Y$, then
\begin{equation}
    f(x) = y
\end{equation}
If an ANN approximating $f$ is called $F$, then we can say that the estimated output of $F$ for the data set $x \in X$ is called $\hat{y} \in \hat{Y}$, such that
\begin{equation}
    F(x) = \hat{y}
\end{equation}
The goal of an ANN, therefore, is to minimise the \textit{loss function} \cite{mlloss} $\mathcal{L}$ over all inputs and outputs
\begin{equation}
    \min_{\theta}\mathcal{L}(\theta) = \min_{\theta} \frac{1}{n} \sum_{i=1}^{n} \mathcal{L}_{\mathrm{loss}}(y, F(x, \theta))
    \label{eq:min}
\end{equation}
where $\theta$ are the trainable parameters that dictate the behaviour of $F$, such as the weights and biases. There are many different ways to measure the loss, (hinge loss, contrast loss, entropy, etc), but the simplest to understand is minimising the square of the euclidean norm of the outputs \cite{loss}, so
\begin{equation}
\begin{aligned}
    \mathcal{L}_{\mathrm{loss}}(y, F(x, \theta)) &= \sqrt{(y_1 - \hat{y}_1)^2 + (y_2 - \hat{y}_2)^2 + ... + (y_n -\hat{y}_n)^2}^2 \\
    & = ||y - \hat{y}||_2^2
\end{aligned}
\end{equation}
meaning that a minimum is found when there is no change that can be made to the parameters $\theta$ that reduces the magnitude of the difference between the function output and the ANN output for the same input.

Clearly, then, finding a minimum set of parameters $\theta$ is somewhat complex. If an ANN has $L$ layers, each with the largest layer having $m$ nodes into $n$ nodes with $n \geq m$, then each pass through the network is approximately $\mathcal{O}(L(nm + n))$. We know $L \leq n$, and so, if $m=n$, a forward pass scales as $\mathcal{O}(n^2)$, as the largest cost is the matrix-vector multiplication (which is completed for every connection). As an example, for the simple ANN described in figure \ref{di:ann}, going from hidden layer 1 to hidden layer 2 requires 42 calculations in of itself, so it is easy to see how this cost can increase dramatically as the size of the ANN increases. 

Instinctively, if the the scenario occurs where $L=n$ then we must only have one neuron in each layer, and thus we have $n$ amounts of scalar multiplications and additions, still leading to $\mathcal{O}(n^2)$.

Training the network is where most of the expense appears. In the worst case scenario, we may require $n$ iterations using $n$ data points to shift $n^2 + n$ weights and biases, giving a time complexity of $\mathcal{O}(n^4)$. This is why it is important to use intelligent strategies when designing and training an ANN.

\section{Optimisation Algorithms}

Today, there exist many methods that one could use to minimise equation \ref{eq:min}. For linear problems, conjugate gradient methods can find the minimum in $n$ steps, where $n$ is the dimension of the input data \cite{conjugate}. Different methods require different knowledge of the function, such has \textit{Jacobians}\footnote{If $x \in \mathbb{R}^n$ are inputs and $f:x\to y\in \mathbb{R}^m$, then the Jacobian $\mathcal{J}\in\mathbb{R}^{m\times n}$ of $f$ is a matrix that holds the partial derivatives of each output of $f_i$ with respect to each input $x_j$, so $\mathcal{J}_{ij}=\frac{\partial f_i}{\partial x_j}$} and/or \textit{Hessians}\footnote{If $x \in \mathbb{R}^n$ are inputs and $f:x\to y\in \mathbb{R}$, then the Hessian $\mathcal{H} \in \mathbb{R}^{n \times n}$ of $f$ is a matrix that holds the 2nd partial derivatives and mixed derivatives of $f$ with respect to each pair inputs $\{x_i, x_j\}_{i,j=1}^n$, so $\mathcal{H}_{ij}=\frac{\partial^2 f}{\partial x_i \partial x_j}$}, but depending on what is known about the function, different strategies can be employed to find a minimum numerically. Some methods even use estimates of the Hessian and/or inverses of matrices, which can speed up computation time \cite{hessest}.

Non-linear problems, such as minimising the loss function for an ANN, are a little more complicated, but numerical methods do exist for these problems (such as Gauss-Newton) \cite{nonlinearreg}. Some methods include \textit{stochastic gradient descent} (SGD), the \textit{limited-memory Broyden–Fletcher–Goldfarb–Shanno} (L-BFGS) algorithm (which is a second order, quasi-newton method, as it uses an estimate of the inverse of the second order derivatives (Hessian) to choose an intelligent search direction), and the ADAM optimiser (which combines two minimisation methods to accelerate learning).

\subsection{Stochastic Gradient Descent}

Normal gradient descent operates by initialising a set of random weights and biases $\theta_0$ and finding the gradient of the loss function across all data points with respect to all parameters for each iteration. We then update $\theta$ by moving in this direction by a magnitude equal to either a pre-selected value, called the training parameter $\gamma$, or a more specific value that requires calculation, called step-size $\alpha$ \cite{overview}.
\begin{equation}
    \theta_{k+1} = \theta_k + \gamma \frac{1}{n}\sum_{i=1}^n\nabla_\theta \mathcal{L}(x_i; \theta_k)
\end{equation}
However, if the data-set $n$ is very large then each step is computationally taxing. To save on this cost, we can divide the data into subsets and minimise randomly over each of these smaller subsets instead, called stochastic gradient descent (SGD) \cite{stochgrad}. This increases iteration speed significantly. There are two main types of SGD
\begin{enumerate}
    \item Full SGD, where each iteration minimises the objective function over an individual data point $x_i \in x$
    \begin{equation}
    \theta_{k+1} = \theta_k + \gamma \nabla_\theta \mathcal{L}(x_i; \theta_k)
\end{equation}
\item Mini-batching, where $n_s \ll n$ number of data points are minimised over per iteration, so
\begin{equation}
    \theta_{k+1} = \theta_k + \gamma \frac{1}{n_s}\sum_{i=1}^{n_s}\nabla_\theta \mathcal{L}(x_i; \theta_k)
\end{equation}
\end{enumerate}

 Mini-batching can be done randomly so that each subset of $x$ is randomly generated per iteration, or the data is pooled into separate subsets before training.

\subsection{ADAM Optimisation}

Adaptive Moment Estimation (ADAM) algorithm combines \textit{adaptive gradident descent} (AdaGrad) with \textit{Root Mean Square Propogation} (RMSProp) \cite{adam}. The main driving force behind Adam is that descent direction takes into account the current momentum \cite{momentum}, and the step size is carefully generated to create a \textit{trust region}, outside of which we cannot be sure what the shape of the objective function is.  ADAM (Algorithm \ref{alg:ADAM}) calculates momentum by using moving averages of the gradient and the squared gradient, with hyper parameters $\beta_1$ and $\beta_2$ controlling the exponential decay.

\begin{figure}[t]
\begin{center}
\includegraphics[width=1\textwidth]{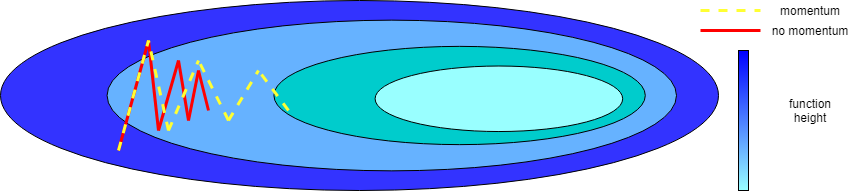}
\caption{Visual representation of how momentum effects descent direction}
\label{di:mom}
\end{center}
\end{figure}

\begin{algorithm}[t]
\caption{ADAM optimisation}\label{alg:ADAM}
\begin{algorithmic}
\Require $\alpha$ \Comment{Step size}
\Require $\epsilon$ \Comment{Threshold for convergence}
\Require $\beta_1, \beta_2\in[0,1)$ \Comment{Exponential decay rates for moment estimates}
\Require $\mathcal{L}$ \Comment{Objective function}
\Require $\theta_0$ \Comment{Initial parameters}
\State $m_0 \gets 0$ \Comment{1st moment vector of 0s}
\State $v_0 \gets 0$ \Comment{2nd moment vector of 0s}
\State $i \gets 0$ \Comment{Set up iteration counter}
\State $\delta \gets 10^{-8}$ \Comment{Prevents division by 0}
\While{$\mathcal{L}(\theta_i) > \epsilon$}
\State $i \gets i+1$ \Comment{Update iteration counter}
\State $\mathcal{J}_i \gets \nabla_{\theta_{i-1}}(\theta_{i-1})$ \Comment{Find gradients}
\State $m_i \gets \beta_1 m_{i-1} + (1-\beta_2)\mathcal{J}_i$ \Comment{Update biased 1st order estimate}
\State $v_i \gets \beta_2v_{i-1} + (1-\beta_1)\mathcal{J}_i \odot \mathcal{J}_{i}$ \Comment{Update biased 2nd moment estimate}
\State $\hat{m}_i \gets \frac{m_i}{1-\beta_1^i}$ \Comment{Bias-corrected 1st order estimate}
\State $\hat{v}_i \gets \frac{v_i}{1-\beta_2^i}$ \Comment{Bias-corrected 2nd order estimate}
\State $\theta_i \gets \theta_{i-1} - \alpha \frac{\hat{m_i}}{\sqrt{\hat{v_i}}+\delta}$  \Comment{Update parameters}
\EndWhile
\State \Return $\theta_i$
\end{algorithmic}
\end{algorithm}
 Usual values for the hyper parameters are $\beta_1=0.9$, and $\beta_2=0.999$, with $a\odot b$ representing the element wise multiplication between vectors $a$ and $b$. The step size is bounded by a region around the current point, outside of which we lack information to know gradients accurately.

\subsection{L-BFGS Optimisation}

For large-scale problems with many parameters, it is important to keep in mind that we cannot operate on unlimited memory. The larger the number of parameters, the larger the Jacobian $\mathcal{J}$ and Hessians $\mathcal{H}$ grow. L-BFGS limits memory usage by only considering the past $m$ updates when estimating the Hessian for the kth step $\mathcal{H}_k$ \cite{lbfgs}.

\begin{algorithm}[t]
\caption{BFGS}\label{alg:BFGS}
\begin{algorithmic}
\Require $\theta_0$ \Comment{Initial guess for optimal parameters}
\Require $\epsilon$ \Comment{Tolerance for minimisation}
\Require $\mathcal{H}_0$ \Comment{First estimate of Hessian (usually $I$)}
\Require $\mathcal{L}$ \Comment{The objective function to minimise}
\State $k \gets 0$
\While{$\mathcal{L}(\theta_k) > \epsilon$}
\State $p_k \gets -\mathcal{H}_k^{-1} \nabla \mathcal{L}(\theta_k)$ \Comment{Find search direction}
\State $\alpha_k \gets \min_{\alpha}\mathcal{L}(\theta_k + \alpha p_k)$ \Comment{Perform line search (exact or inexact)}
\State $\theta_{k+1} \gets \theta_k + \alpha_k p_k$ \Comment{Update parameters}
\State $y_k = \nabla\mathcal{L}(\theta_{k+1}) - \nabla\mathcal{L}(\theta_{k})$ \Comment{Find difference between gradients}
\State $\mathcal{H}_{k+1} \gets \mathcal{H}_k + \frac{y_k y_k^T}{y_k^T s_k} - \frac{\mathcal{H}_ks_ks_k^T\mathcal{H}_k^T}{s_k^T\mathcal{H}_ks_k}$ \Comment{Update Hessian estimate}
\EndWhile
\State \Return $\theta_k$
\end{algorithmic}
\end{algorithm}
The BFGS algorithm (Algorithm \ref{alg:BFGS}) shows that:
\begin{equation}
    \theta_{k+1} = \theta_k + \mathcal{H}_k^{-1} \frac{1}{n}\sum_{k=i}^n \nabla_{\theta}\mathcal{L}(x_i;\theta)
\end{equation}
If $\mathcal{H}_0 = I$, the identity matrix, then the first iteration is equivalent of steepest descent. The inverse of the estimated Hessian can be found directly by considering the \textit{Sherman-Morrison} formula \cite{sher}, so
\begin{equation}
    \mathcal{H}_{k+1}^{-1} = \bigg (I - \frac{s_ky_k^T}{y_k^Ts_k}\bigg)\mathcal{H}_k^{-1}\bigg(I - \frac{y_ks_k^T}{y_k^Ts_k}\bigg) + \frac{s_ks_k^T}{y_k^Ts_k}
\end{equation}
Inexact line search is used to find an $\alpha$ that \textit{sufficiently reduces} $\mathcal{L}(\theta_k + \alpha p_k)$, rather than finding the exact minimum, as this is computationally cheaper. Fundamentally, the learning parameter that is used in gradient descent methods is here replaced with an estimated inverse Hessian, which gives a much more intelligent descent direction. This is because we have approximate knowledge of the gradients and the change in the gradients. If $\mathcal{H}_k^{-1}$ is the exact inverse Hessian, and the algorithmic cost is quadratic, this algorithm can reach the minimum in a single step.

\subsection{Inexact Line Search}

A step size that `sufficiently reduces' the objective function is usually defined in accordance with the \textit{Armijo rule} \cite{armijo} and the \textit{Wolfe condition} \cite{wolfe}.

The Armijo rule states that, given a search direction for the kth iteration $p_k$ has been found, the following inequality must hold
\begin{equation}
    \mathcal{L}(\theta_k + \alpha_kp_k)\leq \mathcal{L}(\theta_k) + c_1\alpha_kp_k^T\nabla\mathcal{L}(\theta_k)
\end{equation}
Where $0<c_1<1$ (usually small, eg $c_1 \approx 10^{-4}$). That is to say that, we expect the step length to update $\theta$ such that it is below a reduced tangent from the point $\theta_k$ in the direction of $p_k$ using back tracking line search. It strikes a balance between
\begin{enumerate}
    \item Not taking steps so large that $\mathcal{L}(\theta_{k+1}) \geq \mathcal{L}(\theta_k)$.
    \item Not taking steps so small that convergence is too long.
\end{enumerate}
Since the algorithm is a backtracking linesearch, it will usually take the maximum value of $\alpha_k$ that satisfies the condition.

The Wolfe condition on curvature states that the following inequality must hold
\begin{equation}
    |p_k^T\nabla\mathcal{L}(\theta_k + \alpha_k p_k)| \leq c_2 |p_k^T \nabla\mathcal{L}(\theta)|
\end{equation}
Where $0 < c_1 < c_2 < 1$ (for quasi-newton methods like BFGS, $c_2 \approx 0.9$). Fundamentally, we expect that if the step we take is minimising the objective function, then the gradient at this point should be less than the previous point (ie, if we are approaching a minimum, the function should be getting `flatter').

\subsection{Validation}

Once an ANN has been trained, it can be tested on a separate labelled data-set to see how well it performs on data it hasn't seen before. This data-set is called a \textit{validation data-set} \cite{val}. The validation phase is incredibly important for a variety of reasons

\begin{enumerate}
    \item The training data-set may not capture every aspect of the problem the ANN is trying to solve. Having a smaller data-set to test the network against can assist a designer in finding gaps in the training data.
    \item It can uncover over fitting, a common issue in ML. The ANN may perform incredibly well on the training data-set, but perform poorly in validation. Methods, such as data augmentation\footnote{Data augmentation is when a designer artificially adds data to a data set by perturbing original data in a measurable way, such as mirroring an image.} and complexity reduction\footnote{Complexity reduction is simply when a designer simplifies the model that is being used, such as removing nodes and layers. In simplest terms, the more nodes and layers a network has, the more degrees of freedom it has. So, by removing some of these degrees of freedom, overfitting can be avoided.}, can then be employed to prevent overfitting.
    \item Though not an issue for PINNs, the validation phase also needs to ensure fairness across protected variables.
\end{enumerate}

\chapter{Neural Networks for Partial Differential Equations}

Classic ANNs and the usual methods of solving or approximating PDEs can suffer from the same complications - a poorly defined problem or a lack of data. For analytical solutions, we need to have closed-form equations for the initial and boundary conditions in the domain of interest. For accurate FEA, we often need a reasonably fine set of data points (usually called a mesh). This mesh needs to be finer still if the PDEs that are being solved require a very smooth solution, as elements need to agree across element boundaries to the kth derivative.

A normal ANN would have some of the same drawbacks. To approximate a complex physical system (to a high tolerance) from data alone may require a dense data set, which is not always available. A classic problem is finding solutions to the 2-dimensional acoustic wave equation, defined as
\begin{equation}
    \rho \nabla \cdot \bigg (\frac{1}{\rho} \nabla u \bigg) - \frac{1}{v^2}\frac{\partial^2 u}{\partial t^2} = -\rho\frac{\partial^2 f}{\partial t^2}
    \label{eq:acou}
\end{equation}
where 
\begin{itemize}
    \item $\rho(x,y)$ is a density function that describes how dense the material is at the point $(x, y)$.
    \item $u(x, y, t)$ is the wave field, the pressure response due to the acoustic wave
    \item $v(x, y) = \sqrt{\frac{k(x, y)}{\rho(x, y)}}$ is the velocity, with $k$ being the adiabatic compression modulus\footnote{Adiabatic compression modulus describes the ability for a material to be squashed. It describes a materials volume decrease as pressure increases.} 
    \item $f(x, y, t)$ is a source term, where force is injected (possibly creating an acoustic wave, such as a small explosion).
\end{itemize}
This equation has applications in earth modelling, and measuring seismic waves (explored in chapter 5). For no source term, and constant density, the equation can be reduced to
\begin{equation}
    \nabla^2u - \frac{1}{v^2}\frac{\partial^2 u}{\partial t^2} = 0
    \label{eq:canwave}
\end{equation}
where $\nabla^2 u = \frac{\partial^2 u}{\partial x_1^2} + \frac{\partial^2 u}{\partial x_2^2} + ... + \frac{\partial^2 u}{\partial x_n^2}$ is the laplacian operator, with each $x_i$ being a spatial variable. This the simplest form of the wave equation, known as canonical form. Unlike numerical methods, using physics-informed neural networks (PINNs) do not rely in discretising the domain in time and space (a process which often introduces errors). Because of this, PINNs have the capability to learn behaviours and outputs outside of the time-space domain, and can learn these behaviours much more quickly.

\section{The Loss Function for PINNs}

Fundamentally, whilst the PDEs themselves do not give us direct information on what the solution should look like, it does give us an indication on how the solution should behave. Using equation \ref{eq:canwave} as an example, the PDE tells us that the solution should be twice differentiable both spatially and temporally, and that the sum of the average pressure in the spatial directions should be equal to the weighted difference in the temporal directions (which is weighted by the velocity squared) at each point. 

If an ANN was given an incredibly dense, well populated, noiseless data-set, it would converge to this behaviour, following from the universal approximation theorem (assuming the size of the network and computational power was no object). However, this level of data cannot always be made available. It is almost impossible, for example, to densely measure the pressure of an acoustic wave over 2km in width and depth. Even if this data set was available, how accurately and precisely could we measure this data? Can we account for the noise, and how will this effect the solution? PINNs work around this lack of data by creating special loss functions that allow the network to converge to a solution by minimising the loss between known outputs and estimated outputs whilst also honouring the expected behaviour of the function described by the PDE. 

Take an ANN $\Lambda$ that estimates the behaviour of a wave field of a function $u$ that is described by equation \ref{eq:acou}. If a standard loss function is described in equation \ref{eq:min}, which we will now call the data loss, and a differential operator that describes acoustic waves $\mathcal{N}$ is defined as
\begin{equation}
    \mathcal{N}(\Lambda) = \rho \nabla \cdot \bigg (\frac{1}{\rho} \nabla \Lambda \bigg) - \frac{1}{v^2}\frac{\partial^2 \Lambda}{\partial t^2} +\rho\frac{\partial^2 f}{\partial t^2}
    \label{eq:physloss}
\end{equation}
Clearly, if $\Lambda$ is an exact solution to the acoustic wave equation, then $\mathcal{N}(\Lambda) = 0$. We can exploit this by adding it as a constraint to the loss function. That is to say, not only do we expect the solution $\Lambda$ to minimise the difference between known outputs $u(x_i, y_i, t_i)$ and estimated outputs $\Lambda(x_i, y_i, t_i; \theta)$, but we also expect the solution to minimise the value of $\mathcal{N}(\Lambda)$. The physics-informed loss therefore looks like \cite{pinnloss}
\begin{equation}
    \mathcal{L}(\theta) = \frac{1-\lambda}{N_d}\sum_{i=1}^{N_d}||u(x_i, y_i, t_i) - \Lambda(x_i, y_i, t_i)||_2^2 + \frac{\lambda}{N_s}\sum_{j=1}^{N_s}||\mathcal{N}(\Lambda(x_j, y_j, t_j))||_2^2
    \label{eq:pinnloss}
\end{equation}
We call the first term the data loss, and the second term the physics loss, where
\begin{itemize}
    \item $ 0 \leq \lambda \leq 1$  is a chosen weight parameter. The larger the value the larger the physics contribution is 
    \item $N_d$ is the size of the data-set
    \item Each $u(x_i, y_i, t_i)$ is a known output of the unknown wave field function $u$ from the data set of size $N_d$
    \item Each $\Lambda(x_i, y_i, t_i)$ is the output of the PINN at points which have known values
    \item $N_s$ is the number of sampling points over the entire domain
    \item Each $\mathcal{N}(\Lambda(x_j, y_j, t_j))$ is an output of \ref{eq:physloss} for the current parameters $\theta$
\end{itemize}

\section{Differentiating a Neural Network}

Clearly then, PINNs rely heavily on accurate derivatives. As shown in chapter 1, Jacobians and Hessians are constructed from gradients (or estimated gradients) of the objective function for a sub-set of data in order to minimise it, and these matrices can be incredibly large. These matrices are then used to adjust the network parameters. If the number of adjustable parameters $\theta$ for a PINN $\Lambda$ is $n$, then $\mathcal{J}\in\mathbb{R}^n$ and $\mathcal{H}\in\mathbb{R}^{n\times n}$, and $n$ can be arbitrarily large. Not only this, but the physics-informed loss from equation \ref{eq:pinnloss} also requires differentiation with respect to the input variables, which could also be of a high dimension. Therefore, it is obvious that an efficient, accurate way of differentiating an ANN is devised.

When computing the derivatives, there are four main options available \cite{adsurv}
\begin{itemize}
    \item \textit{Manual Differentiation}: Taking the known function (in this case, the neural network $\Lambda$) and differentiating it by hand as many times as is necessary.
    \item \textit{Numerical Differentiation}: Using finite difference approximations.
    \item \textit{Symbolic Differentiation}: Using computational expression manipulation libraries, such as SymPy.
    \item \textit{Automatic Differentiation}: Applying the chain rule in sequence to the individual operators in $\Lambda$ to evaluate the derivative.
\end{itemize}

\subsection{Problems with Differential Techniques}

When operating on PINNs, which can have high dimensional inputs and have many parameters, a lot of these techniques present their own set of issues.

\subsubsection{Manual Differentiation}

Whilst, ultimately, this would be the most accurate way to calculate the derivatives, it is an incredibly cumbersome task. Not only are large ANNs complex expressions, but the number of equations one would have to derive could be astronomical (potentially in the millions). For $\Lambda$ this would mean differentiating with respect to each input to the $k$th order to construct the physics loss, with respect to each parameter in $\theta$ to construct the Jacobian, and with respect to each possible pairing of parameters in $\theta$ to construct the Hessian. Of course, humans are also prone to calculation errors.

Not only is the size of the task an issue, but it is also not a dynamic approach at all. Each derivative would have to be coded by hand, and if the design of the network changed in anyway, an engineer would have to restart the entire process from scratch.

\subsubsection{Numerical Differentiation}

Numerical differentiation is a widely used technique, which involves approximating derivatives by taking an estimate from the Taylor series expansion of a function. Take a function $f$. We can estimate the value of the derivative at the point $x$ to an arbitrary precision by taking the Taylor expansion \cite{nderr}
\begin{equation}
    f(x+\Delta x) = f(x) + f'(x)\Delta x + \frac{f''(x)\Delta x^2}{2!} + ... +
    \frac{f^{(n)}(x)\Delta x^n}{n!} + \mathcal{O}(\Delta x^{n})
\end{equation}
If we take $n=1$, the approximate solution of the first derivative can be found, so
\begin{equation}
    f'(x) \approx \frac{f(x+\Delta x) - f(x)}{\Delta x}
\end{equation}
Provided that the step size $\Delta x$ is small enough, the $\Delta x$ term and beyond should be sufficiently small that they can be ignored. However, we also need to ensure that $|\Delta x| > \epsilon$, where $\epsilon$ is machine epsilon\footnote{Machine epsilon is the upper bound on the relative error for floating point (computational arithmetic) numbers. For single precision $\epsilon = 5\times10^{-6}$, for double precision (which is the standard on most modern computers) $\epsilon = 5\times10^{-15}$, and for quad precision (used on specialist machines) $\epsilon = 5\times10^{-33}$} , to ensure that rounding errors are avoided \cite{epsilon}.

As an example, imagine heating up a perfectly square, infinitely thin block of icecream. Take the 2 dimensional heat equation on the $[0,1]\times[0,1]$ square domain with the following conditions
\begin{equation}
    \begin{cases}
      \phi_{t} = \alpha (\phi_{xx} + \phi_{yy}) \quad \textrm{in } \Omega \\
      \phi(0, y, t) = 100 \quad(BC_1) \\
      \phi(1, y, t) = 25 \quad(BC_2) \\
      \phi(x, 0, t) = 200 \quad(BC_3) \\
      \phi(x, 1, t) = 0 \quad(BC_4) \\
      \phi(x, y, 0) = 50 \quad \textrm{in } \Omega \quad(IC)
    \end{cases}
    \label{eq:heatpde}
\end{equation}
with $\alpha = 1.28\times10^{-4}$, which is the thermal diffusivity of the material. The Taylor series can be used to get a finite difference approximation, so
\begin{equation}
\begin{aligned}
    \phi_t &\approx \frac{\phi_{ij}^{(k+1)} - \phi_{ij}^{(k)}}{\Delta t} \\
    \phi_{xx} &\approx \frac{\phi^{(k)}_{i-1j} - 2\phi^{(k)}_{ij} + \phi^{(k)}_{i+1j}}{\Delta x^2} \\
    \phi_{yy} &\approx \frac{\phi^{(k)}_{ij-1} - 2\phi^{(k)}_{ij} + \phi^{(k)}_{ij+1}}{\Delta y^2} \\
\end{aligned}
\end{equation}
Where $k$ is the time step, $i$ and $j$ are the spatial position $x$ and $y$ respectively, $\Delta t$ is the time step, and $\Delta x$ and $\Delta y$ are the spatial steps. If we take $h^2 = \Delta x^2 = \Delta y^2$, then we can make a substitution into the PDE and rearrange to get
\begin{equation}
    \phi_{ij}^{(k+1)} = \bigg(1-\frac{4\Delta t \alpha}{h^2}\bigg) \phi^{(k)}_{ij} + \Delta t \alpha \bigg(\frac{\phi_{ij-1}^{(k)} + \phi_{i-1j}^{(k)} + \phi_{ij+1}^{(k)} + \phi_{i+1j}^{(k)}}{h^2}\bigg)
\end{equation}
That is to say, if we know the entire state of the plate at some time $t=k$, then we can use this information to make an estimate of the heat distribution of the plate at time $t=k+1$.

\begin{figure}[b]
\begin{center}
\includegraphics[width=1\textwidth]{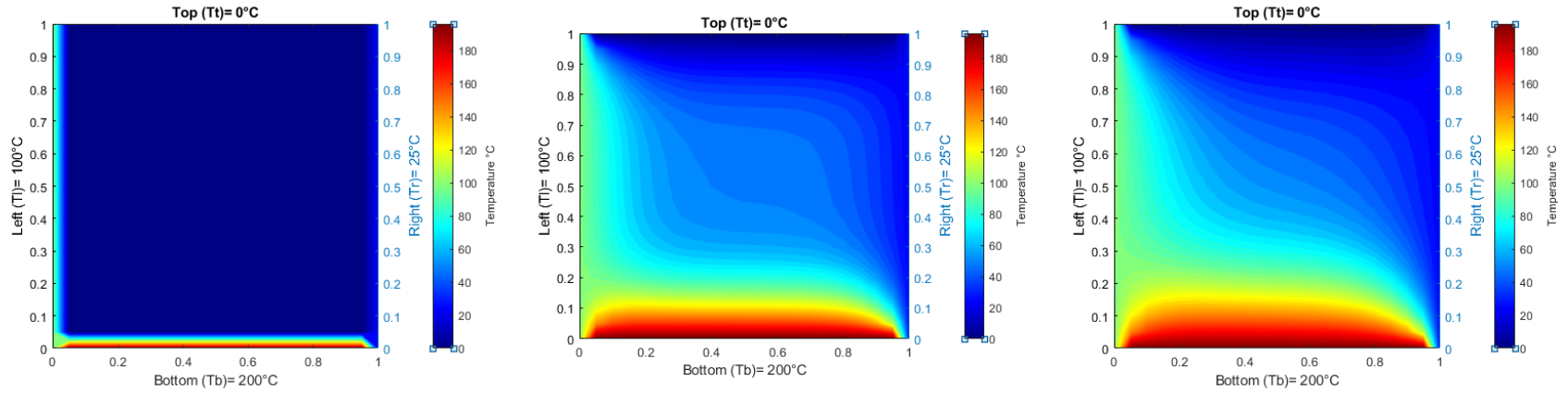}
\caption{Finite difference approximation of the heat equation for \ref{eq:heatpde}, completing the example. From left to right: $t=0$, $t=10$, $t=20$.}
\label{di:temp}
\end{center}
\end{figure}

Clearly, the finite difference method has its merits. It is simple to use and implement, can be quick to calculate for small time and domains, and the error is easy to quantify. It is also reasonably dynamic - changing the structure of the PINN would not alter the code or strategy used to approximate gradients.

It does, also, have its draw backs. One such drawback is that we can only work forward from the ICs \cite{fderr}. If we wanted to know the solution for large $t$, we would need to (accurately) compute all values of $\phi(x, y, t)$ up to this point. Step size in both time and space must also be considered. Too large, and the solution will not be adequately accurate. Too small, and the solution is prone to rounding errors due to machine precision. This problem is exacerbated by much more complex, higher order PDEs, such as fourth order PDEs (sometimes called biharmonic equations)\cite{biharm} which then require that $h^4 > \epsilon$. This sets a real limit on the step size that can be used, and can therefore introduce inaccuracies. Since each iteration's accuracy is also reliant on the previous iteration's accuracy, the errors can easily propagate forward. Fourth order PDEs are not a rare occurrence. A particularly famous one, which appears in structural mechanics and engineering, is the plate bending problem
\begin{equation}
    \nabla^4 w = \frac{p}{D}
\end{equation}
where $p$ is the load distribution on the plate, and $D$ is the Young's modulus\footnote{Young's modulus, named after Thomas Young, is a ratio between the tensile stress strength of the material over the strain strength of the matieral.}. There are PDEs that have an even higher order (that have physical applications), but they are considerably more rare. One 5th order PDE is the Kaup–Kupershmidt equation \cite{kk}, which has similar applications to the more famous Korteweg–De Vries (KdV) equation used in modelling shallow water waves \cite{kdv}.

\begin{figure}[t]
\begin{center}
\includegraphics[width=0.8\textwidth]{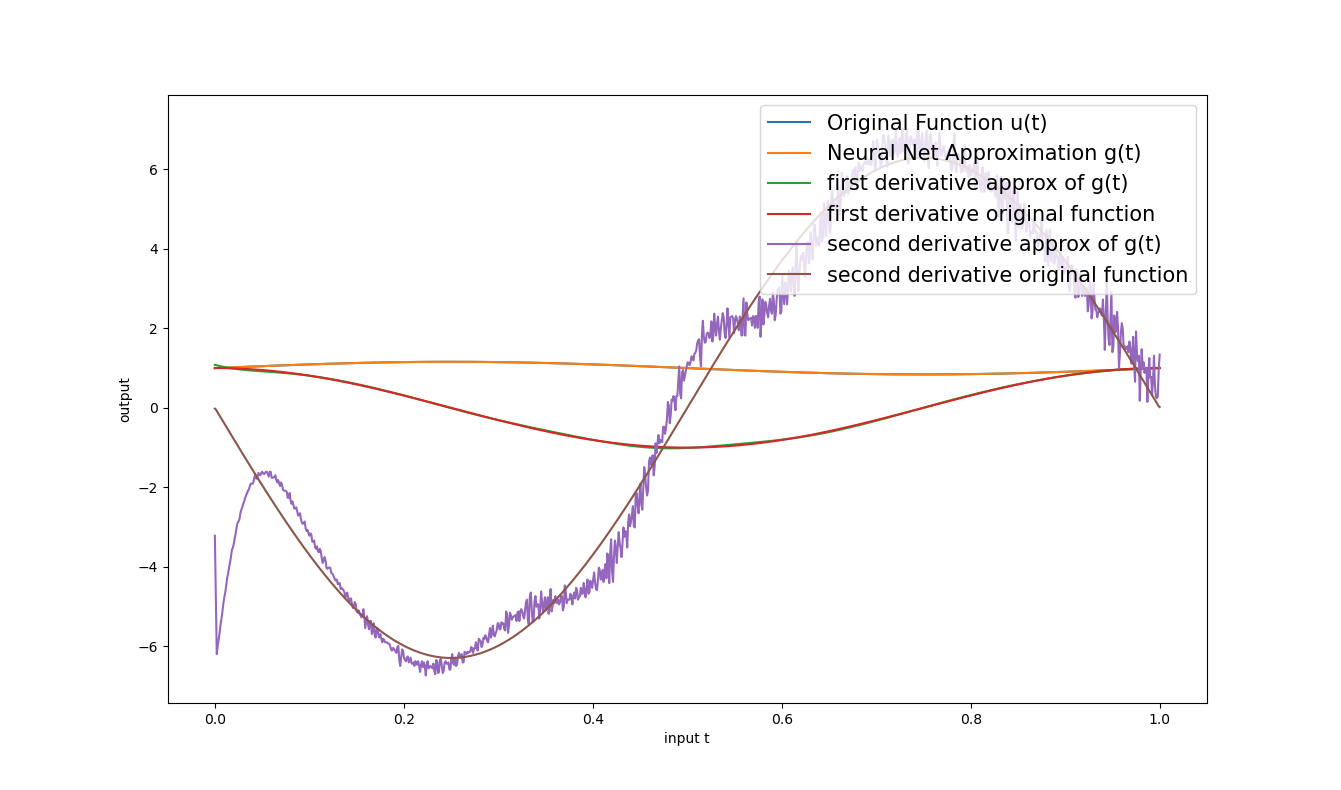}
\caption{A ANN approximation estimating the solution to a simple ODE. The approximate solution is incredibly accurate, as is the 1st derivative. Despite this, estimating the 2nd derivative using finite difference methods shows how much smoothness we can lose when compared to the actual 2nd derivative.}
\label{di:smoothnet}
\end{center}
\end{figure}
 Since machine learning relies on accurate gradients for (sometimes) millions of parameters to optimise the ANN, introducing rounding errors and truncation errors in the training phase can have a huge impact on the ANN's ability to converge to an adequate solution, especially since gradients are already only estimates of the space when using stochastic methods. Badly chosen step sizes can also lead to numerical instabilities. On top of this, when considering a PINN, the loss function includes the PDE as part of the optimisation, and so higher order PDEs would suffer from the same issues as the finite difference method.

\subsubsection{Symbolic Differentiation}

Symbolic differentiation takes the expression described by a function and applies the differential rules (product rule, chain rule, etc) to return a new function $f'$, which is the differential of the function $f$. It is, in truth, very similar to manual differentiation, except that a computer handles the calculation to return a new expression. This is possible because, when it comes to differentiation, there are really only 8 rules to follow - the challenge, therefore, is applying them correctly. 

Symbolic differentiation confronts many of the issues that plague the first two methods. The computer handles the heavy lifting of calculating the expression, which solves the main problems behind applying manual differentiation. Also, the solution that is found is exact, and so we avoid the numerical errors that estimating from the Taylor series expansion bring.

However, it does also have problems when applied to computational functions, especially when applied to ANNs. Firstly, it cannot take into account conditional computational logic (such as if, while, and for). Secondly, the ANN would need to be expressed in a closed form, which, whilst technically possible, adds another challenge. Thirdly, symbolic differentiation is subject to extreme expression swell \cite{adsurv}, (an example of which is in appendix A.1.1). Some rules for derivatives, such as the product rule, naturally lead to an increase in terms in the equation, and so calculating the derivatives can be incredibly cumbersome.

\subsection{Automatic Differentiation}

Automatic differentiation (AD) solves many of the challenges presented above, as it can calculate the derivative at a point to machine precision. It does this by utilising two aspects of compuatational mathematics - \textit{dual numbers} and \textit{computational graph representation} \cite{autodiff}. Fundamentally, all functions are composed of simple operations that can be differentiated, and intermediate variables that store information in the function at different points. AD uses this fact, and exploits the chain rule to combine derivatives of smaller sub functions to find the numerical value of the derivative directly, instead of attempting to calculate a closed form expression, or estimate it using surrounding points.

\subsubsection{Dual Numbers}

Dual numbers are a special way of representing numbers in floating point arithmetic \cite{dual}, which can be leveraged to calculate derivatives of functions at the same time as calculating the primary output. Take $z$ to be a floating point number and $\epsilon$ to be infinitesimally small. We can then say that $z=a+b\epsilon$ , which can be represented in a matrix form as
\begin{equation}
    a = \begin{bmatrix}
            a & 0 \\ 0 & a
    \end{bmatrix}
    \quad
    b\epsilon = 
    \begin{bmatrix}
            0 & b \\ 0 & 0
    \end{bmatrix}
\end{equation}
and so
\begin{equation}
    \epsilon = \begin{bmatrix}
            0 & 1 \\ 0 & 0 
    \end{bmatrix}
    \implies \epsilon^2 = \begin{bmatrix}
                    0 & 0 \\ 0 & 0
            \end{bmatrix}
\end{equation}
Now, take a generic polynomial $P$ of degree $N$, which can be represented in the concise form
\begin{equation}
    P(x) = a_0 + \sum_{n=1}^N a_nx^n
\end{equation}
It follows that
\begin{equation}
    \begin{aligned}
        P(x+\epsilon) &= a_0 + \sum_{n=1}^N a_n(x+\epsilon)^n \\
        &= a_0 + a_1(x+\epsilon) + a_2(x+\epsilon)^2+...+a_N(x+\epsilon)^N \\
        &= a_0 + a_1(x+\epsilon) +a_2(x^2 + 2\epsilon x + \epsilon^2) \\ &+ a_3(x^3 + 3\epsilon x^2 + 3\epsilon^2 x + \epsilon^3) + ...
    \end{aligned}
\end{equation}
However, since $\epsilon^2 = 0$, all terms $\epsilon^k$ with $k \geq 2$ can be removed, and so
\begin{equation}
\begin{aligned}
    P(x+\epsilon) &= a_0 + a_1(x+\epsilon) +a_2(x^2 + 2\epsilon x) \\ &+ a_3(x^3 + 3\epsilon x^2) + ... + a_N(x^N + N\epsilon x^{N-1})\\
    & = a_0 + \sum_{n=1}^N a_nx^n + \epsilon\sum_{n=1}^N a_nnx^{(n-1)} \\
    &= P(x) + \epsilon\frac{\partial P(x)}{\partial x}
\end{aligned}
\end{equation}
So, by evaluating $P(x+\epsilon)$ for a given value $x$ in dual number form, we have calculated the value of $P$ at the point $x$ and the derivative at the point $x$, all in one pass, to machine precision. In general, if a function $f$ is differentiable at a point $x$, then
\begin{equation}
    f(x+\epsilon)=f(x) + \epsilon\frac{\partial f(x)}{\partial x}
\end{equation}
provided that $\epsilon^2$ is computationally 0.

\subsubsection{Computational Graph}

A computational graph is simply a way to represent a function, making it easy to understand the flow of data \cite{graph}. It is very similar to the representation of ANNs in figure \ref{di:ann}, except that it is slightly more generic (as it applies to all types of functions, not just ANNs) and more detailed, as it also shows the operations involved between each step. 

\begin{figure}[b]
\begin{center}
\includegraphics[width=0.65\textwidth]{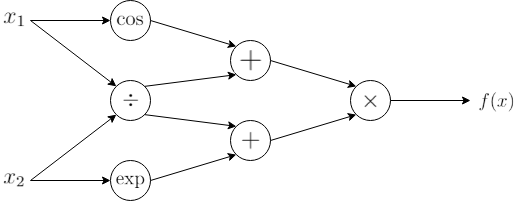}
\caption{Computational graph representation of eq \ref{eq:compgraph}, showing how 2 variables are manipulated together and operated on to generate any output}
\label{di:vibmem}
\end{center}
\end{figure}

A short example would be to imagine a function $f:\mathbb{R}^2\to\mathbb{R}$. If $x\in\mathbb{R}^2$, and
\begin{equation}
    f(x) = \bigg(\frac{x_1}{x_2} + \cos(x_1)\bigg)\bigg(\frac{x_1}{x_2} + e^{x_2}\bigg)
    \label{eq:compgraph}
\end{equation}
then its computational graph representation could look like Figure \ref{di:vibmem}.

This type of representation is useful in three ways. Firstly, it is a reasonably digestible way to see how variables are operated on in a given function. Secondly, it allows the algorithm to exploit repeated operations and variables. As an example, the division $x_1/x_2$ appears twice in equation \ref{eq:compgraph}, and this information is utilised in the graph with the operation only being performed once. Thirdly, this representation allows us to systematically apply the chain on intermediate variables in order to efficiently calculate partial derivatives of the function with respect to the inputs in such a way that it can all be done in one pass.

\subsubsection{Calculating Derivatives}

As mentioned prior, AD does not find an expression for the derivative with respect to any variable, but instead finds the value of the derivative at specific points with respect to any variable. It does this by exploiting the chain rule, which it can do in one of two ways - \textit{forward-mode} AD or \textit{reverse-mode} AD.

Forward AD is more efficient when the function we are differentiating $f$ maps to more outputs then inputs, and vice versa for reverse mode AD. Take the example defined in equation \ref{eq:compgraph}, but this time we want to keep a record of intermediate variable values.

\begin{figure}[b]
\begin{center}
\includegraphics[width=0.85\textwidth]{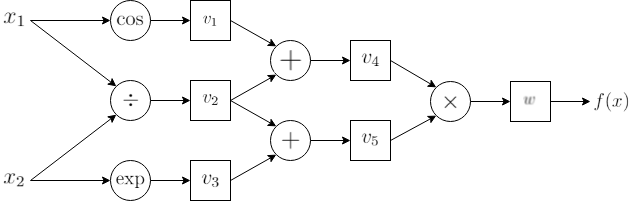}
\caption{Computational graph representation of eq \ref{eq:compgraph}, this time storing the value}
\label{di:compgraphvar}
\end{center}
\end{figure}
As we calculate the values for each variable, (often called the primals), we can also calculate the gradient for each variable with respect to an input (often called the tangents). Assume, for a moment, the input vector is $x=[1, 2]^T$, and we require the value of $f$ at this point, as well as the gradient with respect to $x_1$. Defining $\Dot{v_i}$ as any variable differentiated with respect to $x_1$, we can then perform a forward pass, calculating the primal and tangent of each variable by utilising the properties of dual numbers.

\begin{table}[t]
\begin{center}
\begin{tabular}{@{}r r c r r@{}} 
 \toprule
 \multicolumn{2}{c}{Primals} && \multicolumn{2}{c}{Tangents} \\
 \cmidrule{1-2} \cmidrule{4-5}
 Expression & Value (6 d.p.) && Expression & Value (6 d.p.) \\ %[0.5ex] 
 \midrule
 $x_1 = x_1$ & 1 && $\Dot{x_1} = 1$ & $1$ \\
 $x_2 = x_2$ & 2 && $\Dot{x_2} = 0$ & $0$ \\
 $v_1 = \cos(x_1)$ & $0.540302$ && $\Dot{v_1} = \Dot{x_1}\sin(x_1)$ & $-0.841471$ \\ 
 $v_2 = \frac{x_1}{x_2}$ & $0.5$ && $\Dot{v_2} = x_2^{-1}$ & 0.5 \\
 $v_3 = e^{x_2}$ & $7.389056$ && $\Dot{v_3} = 0$ & 0 \\
 $v_4 = v_1 + v_2$ & $1.040302$ && $\Dot{v_4} = \Dot{v_1} + \Dot{v_2}$ & $-0.341471$ \\
 $v_5 = v_2+v_3$ & $7.889056$ && $\Dot{v_5} = \Dot{v_2} + \Dot{v_3}$ & $0.5$ \\ 
 $w = v_4v_5$ & $8.207001$ && $\Dot{w} = \Dot{v_4}v_5 + v_4\Dot{v_5}$ & $-2.173732$ \\
 %[1ex] 
 \bottomrule
\end{tabular}
\caption{Represented here are the values of the intermediate variables for the computational graph in figure \ref{di:compgraphvar}. By exploiting the chain rule, one forward pass calculates both the value of the function at a given point and its gradient. The computer doesn't really `know' the expression for the tangents - these values are calculated as a byproduct of using dual numbers. They are then used with the chain rule to calculate the derivative.  \label{primalss}}
\end{center}
\end{table}

 Figure \ref{di:compgraphvar} and Table \ref{primalss} show that $\frac{\partial f}{x_1} = \Dot{w} = -2.173732$. We can verify this as true by evaluating the expression of the derivative at the point (which we can do here because $f$ is simple), so
\begin{equation}
    \frac{\partial f}{\partial x_1} = \left(\dfrac{x_1}{x_2}+\mathrm{e}^{x_2}\right)\left(\dfrac{1}{x_2}-\sin\left(x_1\right)\right)+\dfrac{\cos\left(x_1\right)+\frac{x_1}{x_2}}{x_2}
\end{equation}
and so $\frac{\partial f}{\partial x_1}\big|_{x_1=1,x_2=2}$ is -$2.173732$, as expected (in fact, it is correct to 14 decimal places, as we would expect from machine precision). Using a finite difference with a step size of $\Delta x_1 = 1\times 10^{-5}$ yields $\frac{\partial f}{\partial x_1}\big|_{x_1=1,x_2=2}\approx -2.1738$, which is only correct to 3 decimal places, and thus we lose a great deal of precision, even for this very simple function.

Therefore, primals and tangents can be calculated in parallel. Using forward AD, an entire column of the Jacobian is generated in a single pass, and less calculation is required because intermediate values and repeated operations can be exploited more effectively. For $m > n$, this is an efficient way to calculate gradients. However, if $n > m$, we can instead use reverse mode AD to construct the Jacobian each row at a time instead. Constructing the Jacobian column wise means each pass finds the value of all outputs with respect to one input, and constructing the Jacobian row wise finds the value of single output with respect to each input. The cost of running forward mode AD is $\mathcal{O}(n)$, whereas reverse mode is $\mathcal{O}(m)$. Other modes of differentiation either introduce errors, and/or are $\mathcal{O}(nm)$ \cite{eff}.

Many of the functions in physics have a high dimensional input, but a low dimensional output, which means that reverse mode AD is a very good option for calculating derivatives. This is especially apparent in machine learning, where the parameter size is (usually) significantly larger than output size.

\section{The Universal Approximator}

Whilst the main driving force behind PINNs is the way the loss function is defined, there is more to consider when designing the network. Firstly, we should ask ourselves what other information we can directly encode into the PINN to increase the speed of convergence.

Two such pieces of information are the boundary conditions and ICss. Consider, for a moment, the problem of modelling a 2-dimensional vibrating membrane on a $2 \times 3$ rectangular domain. Fix at the edges (\textit{Dirichlet} boundary condition\footnote{Named after \textit{Peter Gustav Lejeune Dirichlet}, the Dirichlet boundary condition is such that the function takes a specific, fixed solution along the boundary of the domain}), and define the initial shape and velocity of the membrane within the domain such that
\begin{equation}
    \begin{cases}
      u_{tt} = 6^2 (u_{xx} + u_{yy}) \quad \textrm{in } \Omega \\
      u=0 \quad \textrm{on } \partial \Omega \quad (BC)\\
      u=xy(2-x)(3-y) \quad \textrm{for } t=0 \quad (IC_1)\\ 
      u_t = 0 \quad \textrm{for } t=0 \quad (IC_2)
    \end{cases}
    \label{eq:vibpde}
\end{equation}
so the domain is $[0,2]\times[0,3]$. This PDE has a series solution
\begin{equation}
\begin{aligned}
        u(x,y,t) = \frac{576}{\pi^2}\sum_{i=1}^\infty \sum_{j=1}^\infty \bigg (\frac{(1+(-1)^{i+1})(1+(-1)^{j+1})}{i^3j^3}\sin\big(\frac{i\pi x}{2}\big)\\\sin\big(\frac{j\pi y}{3}\big)\cos(\pi\sqrt{9i^2+4j^2}t)\bigg)
\end{aligned}
\end{equation}

\begin{figure}[t]
\begin{center}
\includegraphics[width=1\textwidth]{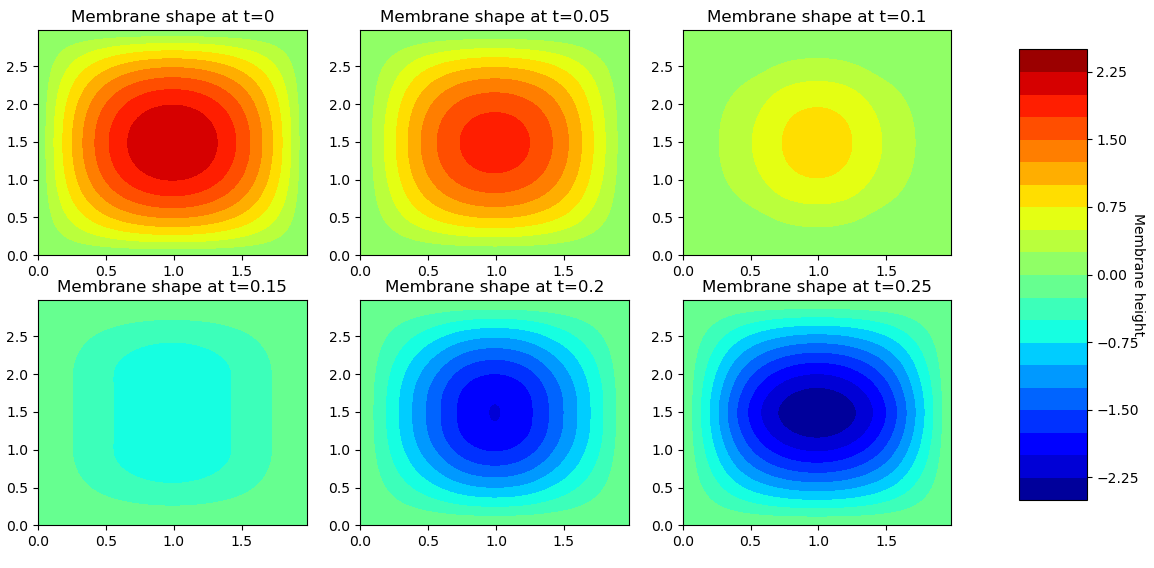}
\caption{Series solution for a simple vibrating membrane with Dirichlet boundary conditions and no dampening, as described in \ref{eq:vibpde} }
\label{di:vibmem3}
\end{center}
\end{figure}

 Take $U$ to be a universal approximator and $\Lambda (x, y, t)$ to be an untrained PINN. In this case, the differential operator in the loss function is simply
\begin{equation}
    \mathcal{N}(\Lambda) = \Lambda_{tt} - 6^2 (\Lambda_{xx} + \Lambda_{yy})
\end{equation}
and so the loss function is defined exactly as it is in equation \ref{eq:pinnloss}, with the new differential operator described above. The first instinct is to simply create the universal approximator $U(x, y, t)$ to solely be equal to the the PINN \cite{badbound}, so
\begin{equation}
    U(x, y, t) = \Lambda(x, y, t)
\end{equation}
However, this means that $\Lambda$ has to learn the ICs and BCs first before applying any time towards learning how to map the rest of the function. Clearly, since this information is already known here, it would be more prudent to give the PINN this information beforehand \cite{goodbound}. To encode the IC, we can instead define
\begin{equation}
\begin{aligned}
    U(x, y, t) &= t\Lambda(x, y, t) + u(x, y, 0) \\
    &=t\Lambda(x, y, t) + xy(2-x)(3-y)
    \end{aligned}
\end{equation}
with the loss function being
\begin{equation}
    \mathcal{L}(\theta) = (1-\lambda)\frac{1}{N_d}\sum_{i=1}^{N_d}||u(x_i, y_i, t_i) - U(x_i, y_i, t_i)||_2^2 + \lambda\frac{1}{N_s}\sum_{j=1}^{N_s}||\mathcal{N}(U(x_j, y_j, t_j))||_2^2
    \label{eq:pinnlossic}
\end{equation}
Now, whenever $t=0$ the PINN automatically satisfies the ICs perfectly, and this behaviour does not need to be approximated at all. We can do exactly the same for the BCs by expanding the universal approximator even further, such that
\begin{equation}
\begin{aligned}
    U(x, y, t) &= u(\partial\Omega, t) t \Lambda(x, y, t) + u(x, y, 0) \\
    &=xy(2-x)(3-y)t\Lambda(x, y, t) + xy(2-x)(3-y)
    \end{aligned}
\end{equation}
Since $f(x, y) = y(2-x)(3-y) = 0$ for $x=0$, $x=2$, $y=0$, or $y=3$ (the extreme parts of the domain of interest), this satisfies the BCs. This small amount of work means the the universal approximator perfectly models the BCs for all time $t \geq 0$, and the ICs for all space within the domain before the training phase has even started.

It is important to note that this is not always possible to do. We may simply lack the knowledge of the BCs/ICs, the data may be noisy or sparse, or the boundaries may be open or ill-defined. However, it does highlight that there are methods that exist outside of the training phase to ensure that the PINN converges to the true solution.

\chapter{Simple PINN Example}

All material in the previous chapters can be applied to build a PINN for almost any problem, provided it is well defined. In cases where the problem is ill-defined, a PINN will still converge to the closest solution it can, and which solution it converges to will often be dependent on the initial parameters, the data it has access to, and the structure of the network. 

\section{General Framework}

This section utilises the framework built by M. Raissi, et al. (2017) \cite{pinnpde}. The paper established that well defined linear and nonlinear PDE problems can be solved to a high level of accuracy using PINNs. Not only this, but a second follow up paper showed that functions and coefficients from a PDE could be reconstructed if enough data in the domain of interest is provided \cite{pdedis}. These two discoveries lay the foundation of the further work that is discussed in the next chapter.

The code (provided in A.4.2, in which there is more detail on use) utilises the Tensorflow package with the L-BFGS optimisation algorithm in Python to approximate solutions to PDEs. In order to produce meaningful results, the program requires 5 things from the user:
\begin{enumerate}
    \item The size of the domain in both spatial and temporal dimensions, so $a \leq x \leq b$ and $0 \leq t \leq T$.
    \item The ICs of the problem for the spatial domain at $t=0$. This can either be as a function, or as discrete data points.
    \item The prescribed BC. Again, this can be delivered as a function for all $0 \leq t \leq T$, or as discrete points.
    \item The network structure (input size, output size, number of hidden layers, and node per layer).
    \item The governing equations to create the custom loss function
\end{enumerate}

 Some other parameters, such as solution precision, max iterations, data sample size, etc can also be changed, though there is a default option. This is because there is a trade off between time to get a solution, and the accuracy that is provided by the PINN.

\section{1D Wave Equation}
\subsection{Analytical Solution}

To demonstrate a simple and intuitive example, take the 1D wave equation with the conditions
\begin{equation}
\begin{aligned}
    \begin{cases}
        u_{tt} = u_{xx} \qquad &\textrm{for } 0 \leq t \leq 4, \quad 0 \leq x \leq 2 \\
        u(x, 0) = x(2-x) \qquad &\textrm{IC} \\
        u_t(x, 0) = 0 \qquad &\textrm{IC} \\
        u(0, t) = u(2, t) = 0 \qquad &\textrm{Boundary Condition}
    \end{cases}
    \end{aligned}
    \label{eq:1dwavenodamp}
\end{equation}
which has the following analytical solution (proof in appendix A.1.2)
\begin{equation}
    u(x, t) = \sum_{n=1}^\infty -\frac{8{\pi}n\sin\left({\pi}n\right)+16\cos\left({\pi}n\right)-16}{{\pi}^3n^3} cos\bigg(\frac{n\pi t}{2}\bigg)\sin\bigg(\frac{n\pi x}{2}\bigg)
\end{equation}

\begin{figure}[t]
\begin{center}
\includegraphics[width=0.8\textwidth]{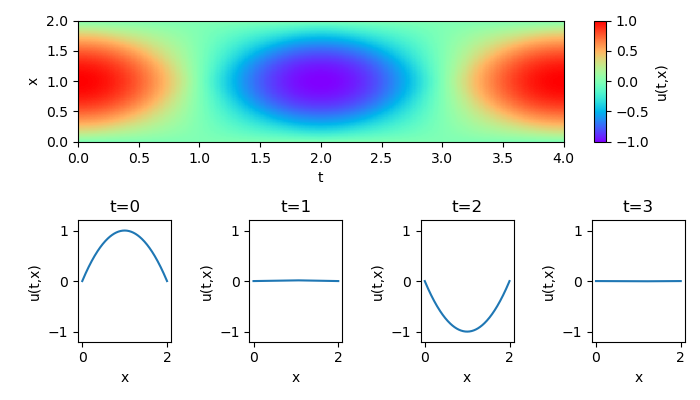}
\caption{Behaviour of the wave described in \ref{eq:1dwavenodamp}. The wave will oscillate like this for all time, since there is no dampening present, and has period 4}
\label{di:1dnodampytrue}
\end{center}
\end{figure}

\subsection{Neural Network}
 In order for a PINN to approximate the solution, no additional information is required at all. In fact, we can approximate a close solution with even less information, as the IC and BC data set might be discrete. In this case, the BC and IC were discretised, and so the network had no knowledge of the behaviour of the IC or BC for all time, only at specific points.

For the 1D wave problem, the PINN $\Lambda(\theta)$ had to minimise
\begin{equation}
    \mathcal{L}(\theta) = \sum_{i=1}^{N_i}\bigg(u(x_i, 0) - \Lambda(x_i, 0; \theta)\bigg) + \sum_{j=1}^{N_b}\bigg(u(x_b, t_j) - \Lambda(x_j, t; \theta)\bigg) + \sum_{k=1}^{N_p}\bigg(\mathcal{N}(\Lambda(x_k, t_k; \theta))\bigg)
\end{equation}
where
\begin{itemize}
    \item The first sum is a loss term for the IC, where $N_i=1000$ are equidistant points taken along the IC
    \item The second sum is a loss term for the boundary, where $N_j=1000$ are randomly selected points along the boundary, so $t_j\in[0, T]$ and $x_b\in\{0, 2\}$
    \item The third term is the physics loss, where $N_p = 10000$ are randomly selected points across the entire domain, and $\mathcal{N}(\Lambda) = \Lambda_{tt} - c^2 \Lambda{xx}$ (with c=1 is the wave speed for this problem)
\end{itemize}
Derivatives are calculated using dual numbers and automatic differentiation by Tensorflow's inbuilt \textit{GradientTape} function, which allows a program to \textit{watch} the network operations with respect to inputs. These functions can be nested in themselves to calculate higher order derivatives.

\begin{figure}[t]
\begin{center}
\includegraphics[width=0.8\textwidth]{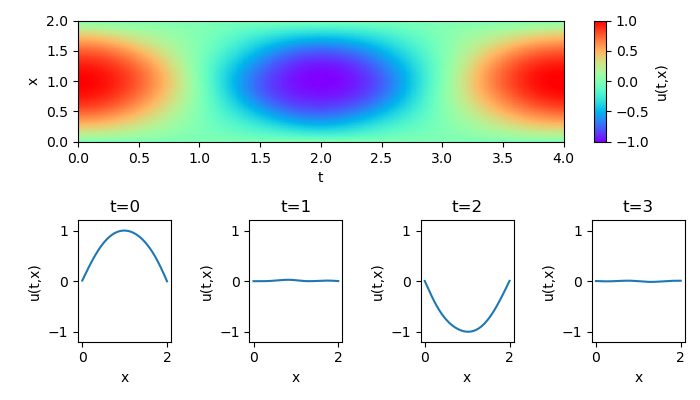}
\caption{Behaviour of the wave as approximated by an PINN. For the above example network structure was $[64, 32, 16, 8]$, so 4 hidden layers with a descending hierarchy.}
\label{di:1dnodampypinn}
\end{center}
\end{figure}

\section{Error Analysis}

The approximate solution provided by the PINN in figure \ref{di:1dnodampypinn} is clearly quite a good fit. The general behaviour is correct, and the solution snapshots at $t=0, 1, 2, 3$ are also very similar to the analytical solution. Since the analytical solution is available, in this case, it is possible to do a reasonably detailed error analysis.

Error analysis of this very simple problem is a crucial endeavor, as it could allow for better informed and more intelligent design considerations when creating PINNs for more complex wave-related problems. It is important to explore where errors occur, in what magnitude they occur, and why they occur.

\subsection{Measuring Error}

Error between the solution and the approximation can be a challenging thing to quantify. However, there are three possible inspirations for estimating the errors between the analytical solution and the PINN solution.

\begin{itemize}
    \item The energy error.
    \item The mean squared error (MSE).
    \item The relative $L_2$ norm error.
\end{itemize}

 Often, we are limited to finding an upper bound on the error, and thus can only know the worst case scenario. FEA estimates the error by studying how the solution differs from the approximation in an \textit{energy norm} sense \cite{energy1} \cite{energy2}.
\begin{equation}
\begin{aligned}
    \int_\Omega(\nabla u - \nabla u_h)\cdot(\nabla u - \nabla u_h) &= ||u-u_h||_E^2 \\
    &=||u||_E^2-||u_h||_E^2
    \end{aligned}
\end{equation}
where $u$ is the exact solution, $u_h$ is the approximate solution, and $\int_\Omega$ is the domain integral. In lieu of this, FEA uses a reference solution to find how the energy norm changes between solutions as the grid is refined, testing for convergence.
\begin{equation}
    ||e_{ref}||_E^2 = ||u_{ref}||_E^2 - ||u_h||_E^2
\end{equation}
where $u_{ref}$ is a fine mesh reference solution, and $u_h$ is a solution on a coarser mesh. This reference error can be bounded by considering the largest possible error, which occurs on the element with the longest edge, or calculated explicitly by running the calculations for a fine mesh. This reference method is worth remembering, as we may not always have access to the true solution $u$. However, in this case we do know the true solution, and can directly use the true energy norm difference. For this specific problem then, the energy norm error is
\begin{equation}
    e_{energy} = \int_0^2\int_0^4 (\nabla u - \nabla \Lambda) \cdot (\nabla u - \nabla \Lambda)
    dtdx
\end{equation}
Since we are dealing with two functions that are continuous in the domain, the energy norm error is a good approximation to use, as other measures require us to discretise the domain.

The \textit{MSE} is defined as
\begin{equation}
    e_{MSE} = \frac{1}{N_e}\sum_{i=1}^{N_e}\bigg(u(x_i, t_i) - \Lambda(x_i, t_i)\bigg)^2
\end{equation}
where $N_e$ is the number of data samples, and pairs $(x_i, t_i)$ are sampled data points. Fundamentally, it is a measure of the average squared error at a finite number of points in the domain. These points can be randomly selected, but here we have taken a defined grid $200\times400$ of equidistant points in $x$ and $t$, giving $N_e = 80000$. 

The final possible method, used in linear algebra, is the \textit{relative $L_2$ norm error}. Again, we take a discrete sample of data points from the solution space and the approximation space and store them as vectors $x$ and $b$ respectively (in this case, 80000 once again). We then find the euclidean distance of the difference between the solution and the approximation vectors, relative to the euclidean distance of the solution.
\begin{equation}
    e_L = \frac{\sqrt{\sum_{i=1}^{N_e}(x_i-b_i)^2}}{\sqrt{\sum_{i=1}^{N_e}x_i^2}} = \frac{||x-b||_2}{||x||_2}
\end{equation}

\subsection{Architecture Design}

Selecting the number of hidden layers and nodes per layer is an important step when designing any kind of ANN. Surprisingly, there is very little solid theory revolving around how and why certain architectures work better than others. When posed with the question `how should I know how many hidden layers to use' in 2013, Yoshua Bengio, who is the Head of the Montreal Institute for Learning Algorithms, stated \textit{`Very simple. Just keep adding layers until the test error does not improve anymore.'}

One school of thought is to study the data-set and see how many straight, connected lines are needed to sufficiently partition the data into defined groups. The number of connections between each line is the number of nodes in hidden layer 1, the number of connections between the set of once-connected lines is the number of nodes in layer 2, etc until all lines are connected \cite{layerline}.

\begin{figure}[t]
\begin{center}
\includegraphics[width=0.8\textwidth]{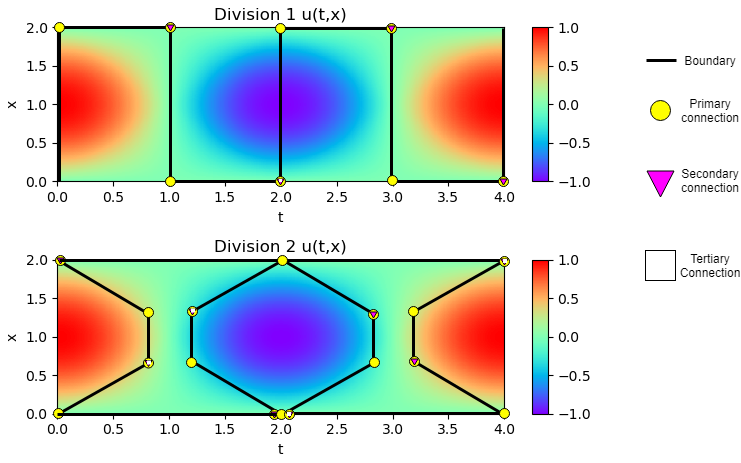}
\caption{Two possible simple partitions of the behaviour of the wave. Division 1 divides the data into sections where derivatives change property, and division 2 categories subsets of data by how close the value is to -1, 0, or 1. Division 1 would imply a 8-4-2 network, whereas division 2 implies a 16-8-4-2 network.}
\label{di:struct_div}
\end{center}
\end{figure}

 Most theories work backwards. That is to say, we create a large ANN that definitely works, and then, through a process called \textit{pruning} \cite{prune}, layers can be removed. A good rule of thumb is to analyse the weights and biases after training. Weights that are approximately 0 (or practically 0 relative to others) can often be removed, as this shows that they have very little impact on the output. Another method is to continually add layers until the error no longer gets smaller, as suggested before. At this point, one layer is then removed to make the function behaviour a little less specific to the training data. Both of these techniques can also prevent overfitting.

Here, some different PINN structures were applied, mainly studying how depth impacted the ability for the PINN to converge.

\begin{table}[t]
\begin{center}
\begin{tabular}{@{}r r r r r@{}} 
 \toprule
 Structure & Parameters & $e_{energy}$ & $e_{MSE}$ & $e_L$ \\ %[0.5ex] 
 \midrule
$128$ & 513 & $8.858\times 10^{-5}$ & $1.956\times10^{-4}$ & $2.7223\times 10^{-2}$ \\ 
 $64-64$ & 4,417 & $3.622\times 10^{-6}$ & $5.509\times10^{-5}$ & $1.444\times 10^{-2}$ \\
 $20-20-20-20$ & 1,341 & $3.522\times 10^{-4}$ & $2.721\times10^{-4}$ & $3.204\times 10^{-2}$ \\
 $32-16-16-32$ & 1,473 & $4.475\times 10^{-5}$ & $1.711\times10^{-4}$ & $2.541\times 10^{-2}$ \\
 $64-32-16-8$ & 2,945 & $1.1815\times 10^{-5}$ & $7.648\times10^{-5}$ & $1.699\times 10^{-2}$ \\
  $8-4-2$ & 73 & $3.10\times 10^{-3}$ & $5.954\times10^{-4}$ & $4.742\times 10^{-2}$ \\ 
   $16-8-4-2$ & 233 & $1.11\times 10^{-3}$ & $5.262\times10^{-4}$ & $4.458\times 10^{-2}$ \\ %[1ex] 
 \bottomrule
\end{tabular}
\caption{Table of errors for differing neural network structures across a spectrum of errors. Most of the architectures give a reasonable estimate of the solution, even when the number of trainable parameter is small. However, training a system with 4417 parameters on an Intel quadcore i7 took over an hour (30000 iterations), whereas 73 parameters took less than a minute. This is why finding sufficiently small network is crucial. \label{params}}
\end{center}
\end{table}

 Upon observing Table \ref{params}, there is clearly a balance to be struck between layer depth and number of nodes. Whilst having the largest number of parameters did create the most accurate approximation, it is not always the case that more parameters means more accuracy. Other much simpler architectures were also reasonably accurate, and there is a huge trade off to consider during the training phase. Training the $[64,64]$ network took over an hour, whereas training the smaller networks, such as the $[8,4,2]$ network, took only minutes. However, the main take away is that even simple networks can capture the general behaviour of a vibrating wave.

\begin{figure}[t]
\begin{center}
\includegraphics[width=1\textwidth]{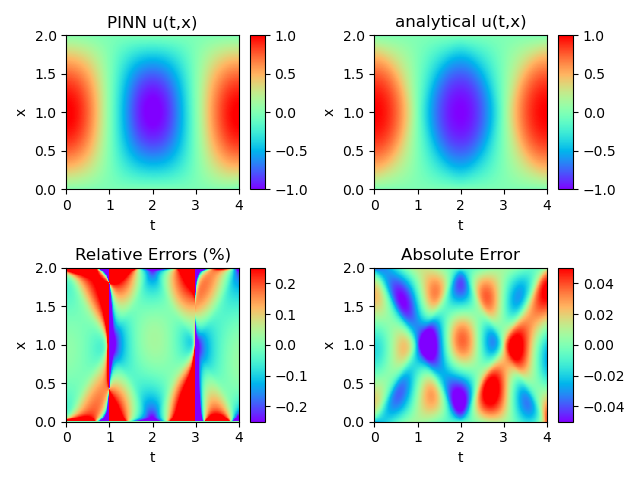}
\caption{Error between true solution and the 8-4-2 network, which was the simplest network tested.}
\label{di:842}
\end{center}
\end{figure}

 Figure \ref{di:842} shows the final solution for the simplest network with the fewest parameters. Whilst the errors shown here are the largest out of any of the networks, it is clear from the plot that the general behaviour is captured. Due to the incredibly short time to train this network, it could be used as a quick estimate to inform a designer how to build a more intelligent network. The absolute time to convergence could be sped up dramatically by using a graphics processing unit (GPU) instead of a CPU, which was unfortunately not an option here. The L-BFGS was used to find a minimum with $m=50$.

\subsection{Solution Convergence}
Knowing the total error for the whole approximate solution is only half of the problem. Something that is more interesting is exploring where these errors occur, and why this might be the case. As an example, let us take the estimate from the $[64, 32, 16, 8]$ network.

\begin{figure}[t]
\begin{center}
\includegraphics[width=0.95\textwidth]{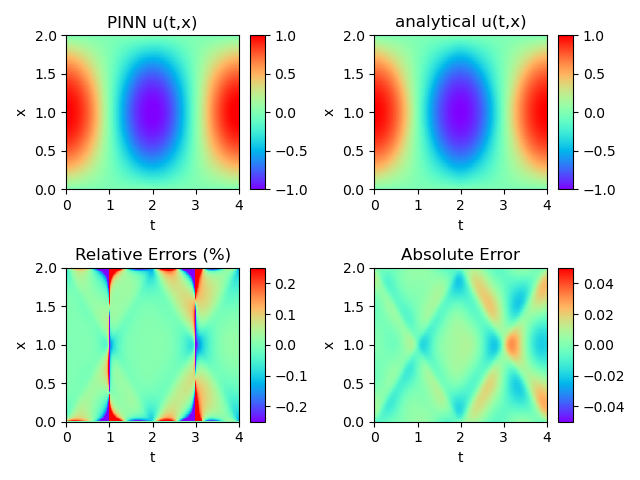}
\caption{Error between true solution and the 64-32-16-8 network}
\label{di:6432168}
\end{center}
\end{figure}

 Judging from table 4.1 and figure \ref{di:6432168}, the solutions is a reasonably good fit. The relative errors are occasionally large, but for good reason. The relative error is calculated as
\begin{equation}
    e_{rel} = \frac{\Lambda(x, t)}{u(x, t)} - 1
\end{equation}
Large relative errors will occur when $u(x, t) \approx 0$ by the nature of the equation used to calculate them. Elsewhere, we can also see that when $|u(x, t)| > 0$ the relative errors are very small, except in the crossing pattern that is noticeable in the absolute errors. For this network, the convergence was tracked. This was done by taking a snapshot of the approximation space every 50 iterations, and the comparing it to the true solution via the $L_2$ error.

\begin{figure}[t]
\begin{center}
\includegraphics[width=0.6\textwidth]{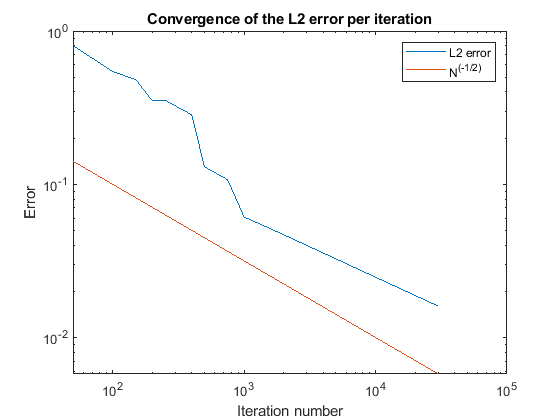}
\caption{Convergence of the $L_2$ error has minimisation progresses}
\label{di:converge}
\end{center}
\end{figure}

The solution approximately converges at a rate of $\mathcal{O}(N^{\frac{1}{2}})$, so 
\begin{equation}
    e_{conv} = \gamma N^{-\frac{1}{2}}
\end{equation}
where $N$ is the number of iterations performed. In this case, $\gamma \approx \frac{5}{2}$. This rate of convergence is also the same for FEA \cite{femconv}, where $N$ would be the number of nodes (also called degrees of freedom) in the mesh. More interesting, however, is how this convergence occurs and where the larger errors are located. 

\begin{figure}[t]
\begin{center}
\includegraphics[width=0.78\textwidth]{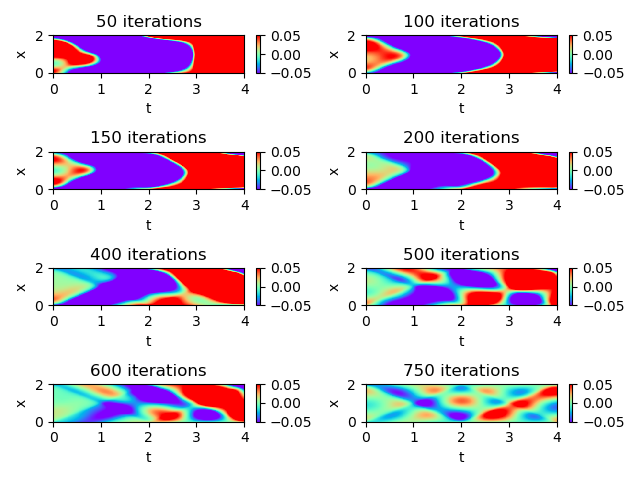}
\caption{Convergence behaviour of $\Lambda(x, t)$ for 64-32-16-8 network. Each figure shows the absolute error against the true solution.}
\label{di:iterations}
\end{center}
\end{figure}

 From Figures \ref{di:842} and \ref{di:6432168}, there seem to be two main observations.
\begin{enumerate}
    \item Errors seem to `propogate' out from the IC, with areas near the IC having a lower absolute error.
    \item The errors have a sort of structure or pattern to them. This structure or pattern is seemingly consistent across differing network architectures (a good example is figure \ref{di:842} vs figure \ref{di:6432168}) - the error is simply more intense on the networks with fewer parameters.
\end{enumerate}

\section{Improving the Learning Process}

The previous section described how the PINN converges to the true solution $u(x, t)$. Knowing this, we can then infer how and why we might improve the performance of the learning process, which could then be applied to much larger scale, more complex problems (which are explored in a later chapter).

\subsection{Error Propogation}

One of the points regarding the error is that, in the approximate solution for the 1D wave problem, errors seem to increase as we move away from the IC, and figure \ref{di:iterations} confirms this.

For this problem, all of the collected data comes from two places
\begin{enumerate}
    \item The boundary of the problem, where $u(0,t)=u(2,t)=0$
    \item The IC of the problem, where $u(x, 0) = x(2-x)$
\end{enumerate}

 Thus, we can infer that to have a low error at $(x=x_1,t=t_1)$, the network must first have a good approximation at $(x=x_n, t < t_1)$, where $x_n$ is a point in the neighbour hood around $x_1$. This conclusion is sensible when we recall that the network learns behaviours by using second order derivatives. In other words, to optimise a single point in space and time it requires good estimates of the surrounding points in space and time, and the best initial estimates will always be near points in the domain that are densely packed with data.

If, instead of taking data from the IC, we take the same amount of data, but at $t=2$ (where $u=x(x-2)$), we should expect the approximation to converge from the centre of the domain moving outwards. 

\begin{figure}[t]
\begin{center}
\includegraphics[width=0.8\textwidth]{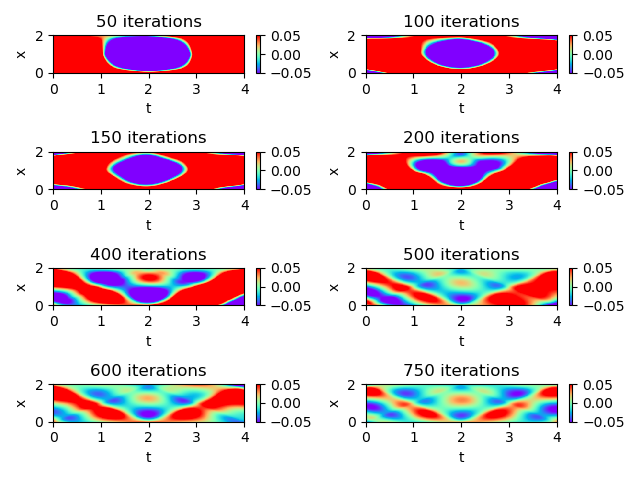}
\caption{Convergence behaviour of $\Lambda(x, t)$ for data taken from $t=2$, rather than taken directly from the IC.}
\label{di:iterations_mid}
\end{center}
\end{figure}

Figure \ref{di:iterations_mid} shows that the prediction appears to be correct. We can clearly see that the optimisation must first find a good solution around the given data before finding the solution closer to $t=0$ and $t=4$. On top of this, the pattern of the errors still seems to take the same shape as for the previous approximation (which used data strictly from $t=0$. This is discussed later.

Most PINN minimisation algorithms, including the algorithm used in this section, take random samples of the domain every iteration to minimise the physics loss term over the entire domain. However, the error propagation highlights that it is somewhat pointless to attempt to learn behaviours for $t= t_0 + t_1$ where $|t_1| \gg 0$ if the behaviour at $t = t_0$ is poorly approximated and $u(x, t)$ is (mostly) known at $t=t_0$. This logic leads to a potential theory in how to converge to a solution in either fewer iterations, or with fewer data points per iteration, thus increasing the computational speed.
\begin{theorem}
    Let $Y$ be the domain in which a PDE is defined. Let $X^{(0)}=X_1^{(0)}\cup X_2^{(0)} \cup...\cup X_{n}^{(0)}\subseteq Y$ where $X_k^{(0)}$ is a part of the domain where known data is dense, partitioned in such a way that $X_i^{(0)} \cap X_j^{(0)} = \varnothing$ for $i \neq j$. Then, for the first $m$ iterations or until a suitable tolerance is found, the loss is defined as
    \begin{equation}
        \mathcal{L}^{(0)}(\theta) = \frac{(1 - \lambda)}{N_d}\sum_{i=1}^{N_d}\bigg(\Lambda(x_i, t_i;\theta) - u(x_i, t_i)\bigg) + \frac{\lambda}{N^{(0)}}\sum_{j=1}^{N^{(0)}}\bigg(\mathcal{N}(\Lambda(x_j^{(0)}, t_j^{(0)};\theta))\bigg) 
    \end{equation}
    where each $(x_i, t_i)$ is a known data point, $(x_j^{(0)}, t_j^{(0)}) \in X^{(0)}$ is a randomly sampled point inside densely packed areas, and $N^{(0)}$ is the number of randomly sampled points from $X^{(0)}$. After a set number of iterations $m$ (or convergence), define $X^{(1)} = X_1^{(1)} \cup X_2^{(1)} \cup ... X_n^{(1)}$  such that each $X_i^{(1)} \supset X_i^{(0)}$ and $X_i^{(1)} \cap X_j^{(1)} = \varnothing$ for $i \neq j$, so $X^{(1)} \supset X^{(0)}$. Define
    \begin{equation}
        \mathcal{L}^{(1)}(\theta) = \frac{(1 - \lambda)}{N_d}\sum_{i=1}^{N_d}\bigg(\Lambda(x_i, t_i;\theta) - u(x_i, t_i)\bigg) + \frac{\lambda}{N^{(1)}}\sum_{j=1}^{N^{(1)}}\bigg(\mathcal{N}(\Lambda(x_j^{(1)}, t_j^{(1)};\theta))\bigg) 
    \end{equation}
    where each $(x_j^{(1)}, t_j^{(1)}) \in X^{(1)}$ is a randomly sampled point inside $X^{(1)}$, and run until convergence or $m$ iterations. Continue until $X^{(k-1)} \subset X^{(k)}=X_1^{(k)}\cup X_2^{(k)} \cup...\cup X_{n}^{(k)} = Y$, where $X_i^{(k)} \supset X_i^{(k-1)}$ and $X_i^{(k)} \cap X_j^{(k)} = \varnothing$ for $i \neq j$. Then, use the usual loss function that takes random samples from the whole domain.
    \begin{equation}
        \mathcal{L}^{(k)}(\theta) = \frac{(1 - \lambda)}{N_d}\sum_{i=1}^{N_d}\bigg(\Lambda(x_i, t_i;\theta) - u(x_i, t_i)\bigg) + \frac{\lambda}{N^{(k)}}\sum_{j=1}^{N^{(k)}}\bigg(\mathcal{N}(\Lambda(x_j^{(k)}, t_j^{(k)};\theta))\bigg) 
    \end{equation}
\end{theorem}

\begin{figure}[t]
\begin{center}
\includegraphics[width=0.8\textwidth]{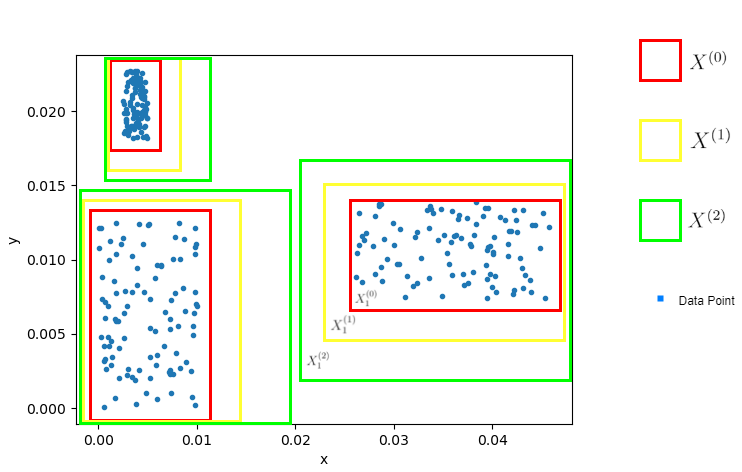}
\caption{One such possible grouping of data for a data-set over a 2D domain. Each iteration of $X^{(i)}$ grows the accessible domain around each group of data. The algorithm would model behaviours inside each partition before expanding to behaviours in other parts of the domain}
\label{di:partition1}
\end{center}
\end{figure}

 Theorem 1 suggests that the optimisation process should focus on approximating the function behaviour near data points before expanding the domain and resolving errors in places with little data. This is because areas that lack data rely on the accuracy of surrounding behaviours to model the solution correctly, so minimising errors in dense data areas should be the initial priority.

\subsection{Error Patterns}

The second point highlights that the absolute error is not randomly distributed through the domain, but appears to have structure. This structure looks to be consistent in each approximate solution, and is unaffected by the location of the data or the network structure.

\begin{figure}[t]
\begin{center}
\includegraphics[width=0.8\textwidth]{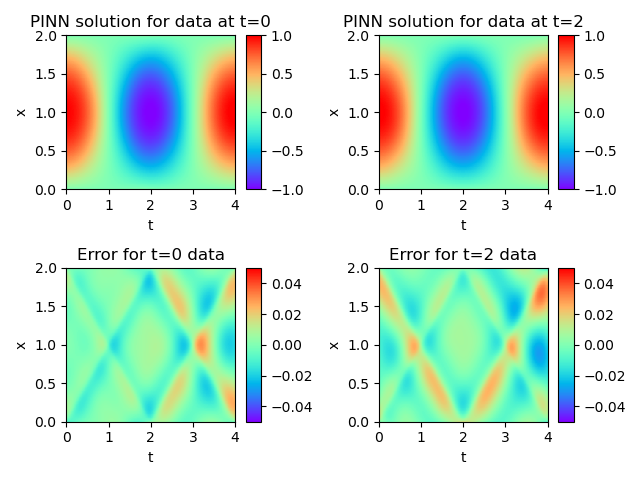}
\caption{The approximate solution and error patterns for data sets from different domain locations. The pattern is independent of the data distribution}
\label{di:partition2}
\end{center}
\end{figure}

 Since the physical behaviour of waves is reasonably intuitive to understand, it is easy to see that these errors seem to occur when the wave is at its `curviest'. The errors seem to swap in a crossing pattern, going from the boundary, to the centre, and back. This is likely because of errors in the derivative.
\begin{itemize}
    \item Errors appear at the boundary when $|u(x, t)| \to 1$ because this is when $\big|\frac{\partial u}{\partial x}\big|$ is at its largest (the wave exhibits maximum spatial curvature).
    \item Errors appear in the centre of the wave when $|u(x, t)| \to 0$ because this is when $\big|\frac{\partial u}{\partial t}\big|$ is at its largest (the centre of the string is experiencing large relative speed, or maximum temporal curvature).
\end{itemize}

\begin{figure}[t]
\begin{center}
\includegraphics[width=0.8\textwidth]{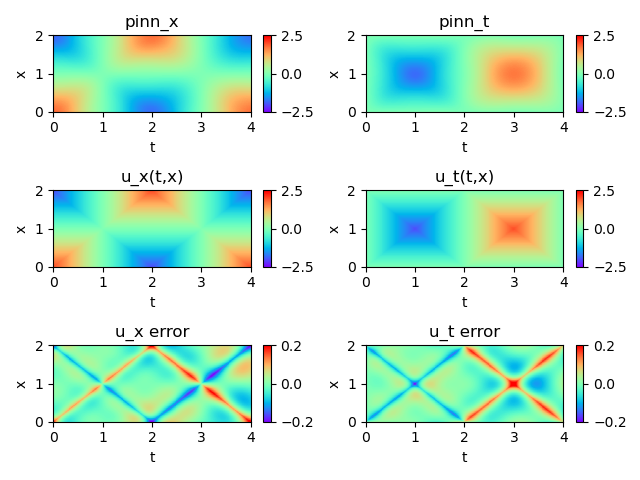}
\caption{Errors between the analytical solution derivatives and the PINN solution derivatives for all data at $t=0$}
\label{di:partition3}
\end{center}
\end{figure}

 Errors in the derivative obviously follow the pattern seen in the approximate solution error. The errors also appear be of a higher relative magnitude, and (as before) it propagates away from the IC, with the best estimate of the overall derivative being where data is most dense.

This discovery is somewhat interesting, as it appears that analysing where rapid change occurs in the PINN could tell us information about where errors may appear without having to know the analytical solution. As a litmus test, the same analysis was performed on a quasi-linear PDE.

Quasi-linear PDEs are a special subclass of PDE in that they can develop \textit{shock formations}\footnote{Shocks form in solutions at $t_s$ for PDEs when the point $u(x, t_s)$ has conflicting information on what value it should take for some $x$ (ie, the solution becomes multi-varied for $t\geq t_s$). This often implies that the assumptions of the governing equations are not valid beyond this point}. As the solution tends towards a shock point, we often find very sharp changes in behaviour, or even discontinuities. 

A famous example of a quasi-linear PDE is the viscous Burger's equation, which is used in modelling fluid mechanics, nonlinear acoustics, and gas dynamics. It is defined as

\begin{equation}
    \phi_t + \phi \phi_x = \nu u_xx 
\end{equation}
where $\nu$ is the viscosity parameter. For small values of $\nu$, the solution can develop shocks, and these can make the system very challenging to solve. Take the following problem, described and solved in Raissi, M. et al. \cite{pinnpde}
\begin{equation}
\begin{aligned}
    \begin{cases}
        \phi_t + \phi \phi_x - \frac{0.01}{\pi}\phi_{xx} = 0 \qquad &\textrm{for } 0 \leq t \leq 1, \quad -1 \leq x \leq 1 \\
        \phi(x, 0) = -\sin(\pi x) \qquad &\textrm{IC} \\
        \phi(-1, t) = \phi(1, t) = 0 \qquad &\textrm{BCs}
    \end{cases}
    \end{aligned}
\end{equation}

\begin{figure}[t]
\begin{center}
\includegraphics[width=0.8\textwidth]{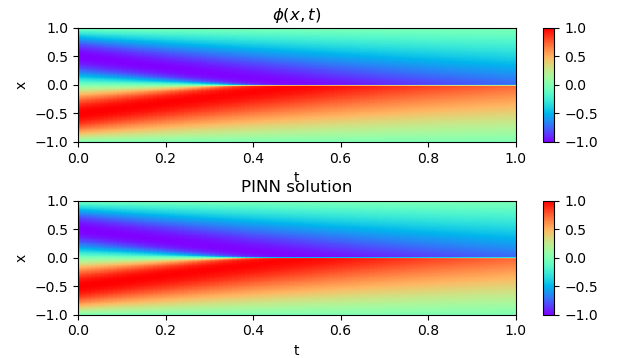}
\caption{True solution to the Burger's equation and the PINN approximate solution}
\label{di:burgers}
\end{center}
\end{figure}

 The data for the solution is taken directly from the cited paper. Clearly, the approximate solution looks to be a good fit, and the $L_2$ error was $4.9\times 10^{-4}$. However, we are interested in where errors appear, if they do at all.

From looking at the form of the solution in figure \ref{di:burgers}, and following on from the logic described in the analysis of error locations for the 1D wave equation, we should expect 2 areas of error.
\begin{enumerate}
    \item Small errors approximately for $t<0.4$ in a delta shape (pointing right) due to the derivative with respect to time. The $t$ derivative is largest here, but it isn't enormous.
    \item Large errors for approximately $t>0.4$, $x\approx 0$ due to the large derivative with respect to space. In fact, it almost becomes a discontinuity.
\end{enumerate}

\begin{figure}[t]
\begin{center}
\includegraphics[width=1\textwidth]{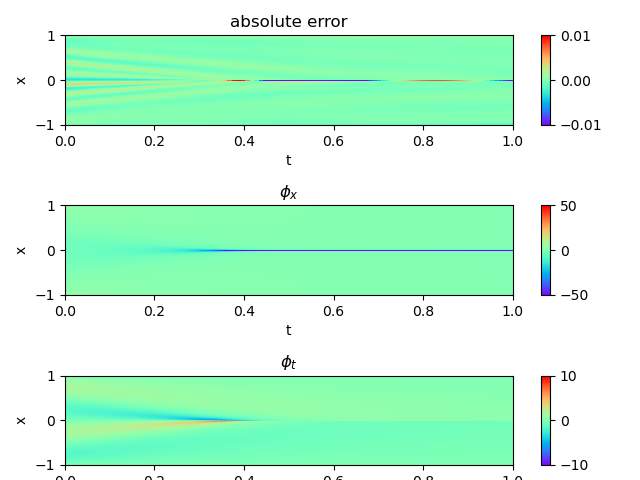}
\caption{Top: The error between the analytical solution and the PINN approximate solution. Middle: the derivative with respect to space. The derivative at $x \approx 0$ dwarfs all others. Bottom: the derivative with respect to time. It is significantly smaller than the extremes in the spatial derivative}
\label{di:burgers_error}
\end{center}
\end{figure}

 From Figure \ref{di:burgers_error}, the prediction was clearly accurate. Despite the small $L_2$ error documented in the study, the error is almost entirely caused by the incredibly steep derivatives in the $x$ direction. This reasons that we can improve the approximate solution with more intelligent data sampling.

\begin{theorem}
    Take $\Lambda(x, t)$ to be a physics-informed neural network which closely approximates the general behaviour of some function $u(x, t)$, and take $\epsilon > 0$ to be the acceptable tolerance to say the solution has converged. Take $m$ to be the maximum number of iterations. If, on iteration $m$, we have
    \begin{equation}
        \frac{1-\lambda}{N_d}\sum_{i=1}^{N_d}||u(x_i, y_i, t_i) - \Lambda(x_i, y_i, t_i)||_2^2 + \frac{\lambda}{N_s}\sum_{j=1}^{N_s}||\mathcal{N}(\Lambda(x_j, y_j, t_j))||_2^2 > \epsilon
    \end{equation}
    then the network has failed to converge to a solution in finite time. Take $\Lambda(x, t)$ and calculate
    \begin{equation}
        \mathcal{J}_\Lambda = \nabla \Lambda(x, t)
    \end{equation}
    such that the Jacobian $\mathcal{J}_\Lambda$ is the vector that holds the derivative of $\Lambda(x, t)$ with respect to each variable. Since errors are most likely to occur when $\mathcal{J}_\Lambda$ is relatively large, we can use $\mathcal{J}$ as a probability density function for data sampling, instead of sampling uniformly across the domain. That is to say
    as
    \begin{equation}
        \frac{||\nabla \Lambda(x_i, t_i)||_2}{\max ||\nabla \Lambda(x, t)||_2} \to 1, \quad P[(x_i, t_i) \in N_s] \to 1
    \end{equation}
    where $N_s$ is the physics term sampling data-set.
\end{theorem}

Fundamentally, if we suspect that, after training, the PINN approximation may have large errors due to derivatives we can introduce another training step (or, perhaps, it could be referred to as a PINN validation step). Due to the fact that errors are likely to be occurring in areas of high relative change (large derivatives), in late training phases we should sample from these areas more densely, and then check if the PINN is minimised over the whole domain. 

The suggestion here is to use the derivatives in the domain as a surrogate for a probability density distribution. The higher the derivative is in a certain area of the domain, the more likely we are to sample a point from this location.

\chapter{Current Applications and Research Areas}

PINNs have already been put to use in academic settings in order to produce neural networks that solve/estimate physical systems. These problems range from small scale problems, like MRI scans \cite{mri} and natural language processing\footnote{NLP is the computational technique of taking in continuous data (such as wavelengths, amplitudes, etc) and analysing the data in such a way that a computer can and interpret individual words and speech} \cite{nlp}, to large scale problems, such seismic imaging and fluid dynamics.

PINNs can be utilised to solve the forward problem of estimating the solution to a PDE, or the inverse problem of approximating a function coefficient in a system of equations. Both types of problem will be explored here.

\section{Geophysical Tomography}

\textit{Geophysical tomography} is the process of applying non-destructive strategies to investigate the properties and structures of subsurface terrain \cite{geotom}. Many different types of techniques can be used to model sub-surface structures, depending on the size of the domain and the detail a geologist may need. 

One widespread application, which can produce high resolution images, is seismic imaging, which uses strong acoustic waves measured at different points to construct an approximation of underground parameters. \textit{Wavefield reconstruction inversion} (WRI) is one such method that is particularly popular \cite{wri1}.

\subsection{Wavefield Reconstruction Inversion}

WRI is an alternative approach to seismic imaging, and builds upon techniques established in \textit{forward wavefield inversion} (FWI). FWI is a data-driven, constrained, nonlinear optimisation method that uses the known physics of the acoustic wave equation to reconstruct features from partial measurements of the wave equation, specifically in frequency-domain, rather than time-domain \cite{wri2}. The preference of using frequency domain over time domain is that waves of different frequencies behave differently depending on the properties of the medium it is passing through. They can reflect, refract, sheer, and penetrate. The frequency domain wave (Helmholtz) equation is
\begin{equation}
    (\nabla^2  + \frac{\omega^2}{c(x)^2})u(x, \omega)=q(\omega)\delta(x-x_s)
\end{equation}
where $u$ is the acoustic seismic wavefield in frequency domain, $\omega$ is the angular frequency, $x$ is the spatial location, $c$ is the velocity of the medium, and $q(\omega)\delta(x-x_s)$ is the frequency domain source at $x=x_s$ (so $\int_{-\infty}^{\infty}q(\omega)\delta(x-x_s)dx = q(\omega)$). In short hand, the discrete Helmholtz operator $A$ (also known as the impedance matrix) will be such that
\begin{equation}
    A(m, \omega) = L^2  + \omega^2\textrm{diag}(m)
\end{equation}
where $L^2$ is the discretised Laplacian operator $\nabla^2$, and $m$ is the discrete squared slowness of the medium $\frac{1}{c(x)^2}$.

The basic premise is that acoustic waves are fired into the surface, and the wavefield response is measured by remote sensors, often on or near the surface of the terrain. Each sensor $d_r$ collects data points from each acoustic source $q_s$.

\begin{figure}[t]
\begin{center}
\includegraphics[width=0.8\textwidth]{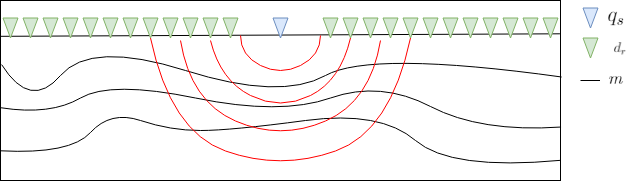}
\caption{A basic representation of imaging using seismic waves. Waves (in red) are propelled through the medium $m$ (in most cases, earth) from a source $q_s$, and sensors $d_r$ take measurements of the wave response on the surface. The response changes depending on the structure of the medium}
\label{di:FWI}
\end{center}
\end{figure}
Using this data, a constrained optimisation problem can be obtained, such that
\begin{equation}
    \min_{m, u} \frac{1}{2}\sum_\omega\sum_{s, r}||\Gamma_{s, r}u_s(\omega) - d_{s, r}||^2_2 W_{s,r} \qquad \textrm{s.t. }A(m, \omega)u_s(\omega) = q_s(\omega)
\end{equation}
Where $W_{s, r}$ is a weight operator applied to the data residual. The constraint can be eliminated by using inner products and Lagrangian multipliers, called the \textit{adjoint-state} or \textit{reduced-space} formulation such that
\begin{equation}
    \mathcal{L}(m, u_s, v_s) = \frac{1}{2} \sum_{s, r}||\Gamma_{s, r}u_s - d_{s, r}||^2_2 W_{s,r} + \sum_s\langle v_s, A(m)u_s- q_s\rangle_x
\end{equation}
where $\langle\cdot,\cdot\rangle_x$ is the inner product on spatial coordinates, and $v_s$ is the Lagrange multiplier. Taking the partial derivative with respect to each minimisation parameter yields
\begin{equation}
    \begin{aligned}
        \frac{\partial \mathcal{L}(m, u_s, v_s)}{\partial m} &= \sum_s \bigg( \frac{\partial A(m)u_s}{\partial m}\bigg)^*v_s \\
        \frac{\mathcal{L}(m, u_s, v_s)}{\partial u_s} &= \sum_{s,r}\Gamma_{s,r}^*W_{s,r}^*(\Gamma_{s, r}u_s - d_{s, r}) + \sum_s A^*(m)v_s \\
        \frac{\partial \mathcal{L}(m, u_s, v_s)}{\partial v_s} &= \sum_s A(m)u_s - q_s 
    \end{aligned}
\end{equation}
where $*$ is the conjugate transpose. The last equation implies that if we are at a minimum then $A(m)u_s=q_s$, and therefore the constraint in the original minimisation problem is satisfied. To be concise, solving the wave equation and the adjoint equation each iteration yields the gradient, which can then be used to calculate a local minimum.

However, this type of inversion is plagued by three main problems:
\begin{enumerate}
    \item Solving the wave equation for all space twice per iteration is computationally taxing, and the constructed Hessian is usually dense.
    \item There is an extreme non-linear dependence on the earth model $m$
    \item The problem is nonconvex, and so solutions are very rarely unique. Thus, the optimal solution found is dependent on the starting parameters of the earth model $m$. In other words, local minima are real complication. This is often called cycle skipping, and occurs when the data generated by the model $\Gamma_{s,r}u_s$ is more than half a cycle away from the recorded data $d_{s,r}$.
\end{enumerate}

 WRI tackles this problem by giving the objective function more degrees of freedom. It achieves this by relaxing the physics constraint slightly by some weight $\lambda^2$ \cite{wri1}, and is instead added directly into the objective function as a penalty parameter.
\begin{equation}
    \min_{m,u_s}\mathcal{L}_\lambda(m,u_s) = \sum_{s,r}\big(||\Gamma_{s,r} u_s - d_{s, r}||^2_2 + \lambda^2||A(m)u_s-q_s||^2_2\big)
\end{equation}
In other words, we expect the optimal earth solution to minimise a balance between the physics and the data, rather than perfectly conforming to the physics. This removes the need to solve the adjoint wave equation, and the Hessian is sparse if $\lambda$ is small (with small $\lambda$ being interpreted as not relying on the physics as much) \cite{wri2}. Recently, WRI has been formulated in time-domain problems, rather than frequency domain, to alleviate some computational cost (as LU factorisation can be more easily leveraged to solve the many linear systems) \cite{wrilu}.

\subsection{Solving the Wave Equation with PINNs}

The two cornerstones of wavefield inversion are
\begin{enumerate}
    \item Constructing a wavefield that satisfies the source data and the receiver data.
    \item Using the estimated wavefield to reconstruct the terrain under the surface boundary
\end{enumerate}

 Physics-informed machine learning has shown great promise in accurately reconstructing wavefields for complex domains, as shown by the University of Oxford in \textit{Solving the Wave Equation with Physics-informed Deep Learning} (B. Mosely et al.) \cite{forwardwave}. The paper applies the methods and strategies highlighted in chapters 2 and 3 to estimate the acoustic response to waves propagated through mediums of varying complexity.

\subsubsection{Method}

The study used the 2D acoustic wave equation (as described in equation \ref{eq:acou}) to create a loss function
\begin{equation}
    \mathcal{L}=\frac{1}{N_u}\sum_{i=1}^{N_u}||u(t_i, x_i,s_i) - \Lambda(t_i, x_i,s_i)||^2 + \frac{1}{N_\Lambda}\sum_{j=1}^{n_\Lambda}||\mathcal{N}(\Lambda(t_j,x_j,s_j;\theta);\lambda)||^2
\end{equation}
where $(t_i, x_i)$ are known initial values, $(t_j, x_j)$ are points sampled inside the whole domain, $\Lambda$ is the PINN, $\lambda$ is a weight function for the physics minimisation, $\theta$ are the trainable parameters that define $\Lambda$, and $\mathcal{N}$ is the differential operator that describes the acoustic wave equation, as defined in equation \ref{eq:physloss}. Surprisingly, AD was not used in order to accurately calculate the gradients needed for the physics term in the loss function, but the authors do note that it could (and, really, should) be used.

Network architecture, and how different architectures may change the solution, was not a variable that was explored in this study. Motivated by Raissi et al. (2019), the activation function $\sigma:\mathbb{R} \to [0,\infty]$ used was the SoftPlus activation function
\begin{equation}
    \sigma(x) = \ln(1+e^x)
\end{equation}
and has a derivative equal to the logistic map, $\sigma'(x) = \frac{1}{1+e^{-x}}$. It is often used because it has a reduced likelihood of vanishing gradients. The PINN design was fixed at 10 layers with 1024 hidden channels, and was fully connected. The input space was the spatial coordinates $x\in\mathbb{R}^2$, the temporal coordinate $t\in\mathbb{R}$, and the source term $s$, with the output being the wavefield $u\in\mathbb{R}$. Each layer was put through the activation function, except the final layer, which was just kept as linear (otherwise the network output would always be $\Lambda(x, t) > 0$).

\begin{figure}[t]
\begin{center}
\includegraphics[width=0.8\textwidth]{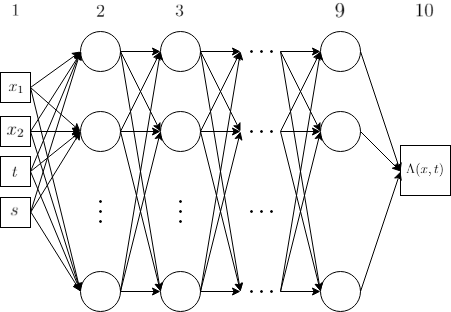}
\caption{Network architecture for solving the wave equation. 10 layers make up the network, with a total of 1024 nodes. The final output layer was linear, rather than using a non-linear activation function.}
\label{di:wavepinn}
\end{center}
\end{figure}

 The study used the ADAM optimisation algorithm to calculate a minimum for the network weights and biases.

\subsubsection{Data}

There were two types of data needed for this study: \textit{initial wave} data and \textit{medium velcoity} data. Since the study was only concerned with the forward problem of modelling the wavefield, the medium density and velocity was known prior. The study used 3 data sets of varying complexity.

\begin{figure}[b]
\begin{center}
\includegraphics[width=0.8\textwidth]{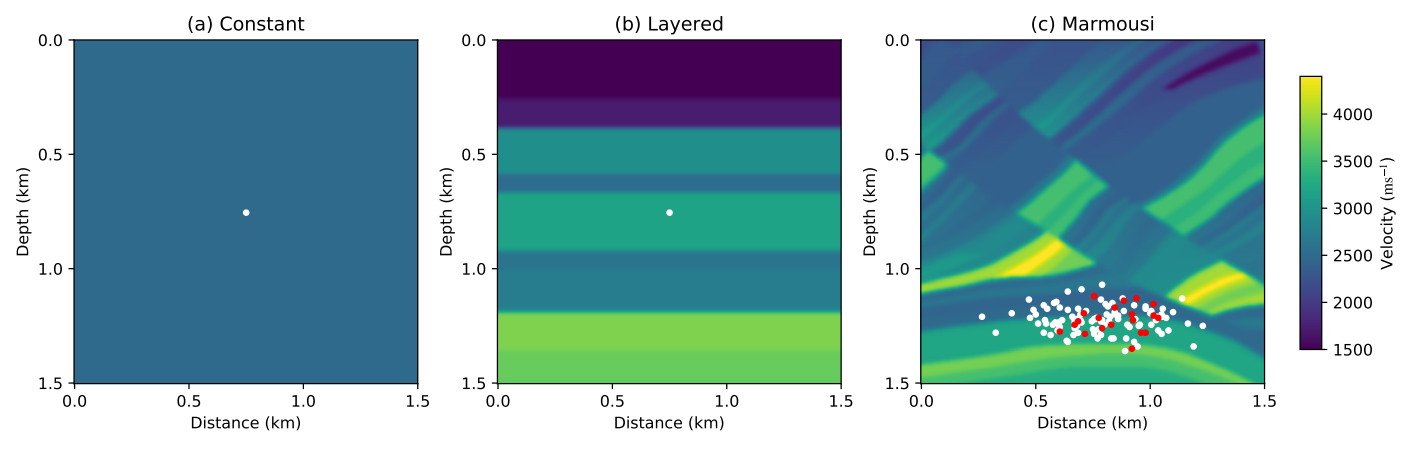}
\caption{Different data sets describing the medium velocity. White points are sources that generate acoustic waves}
\label{di:medvel}
\end{center}
\end{figure}

 Data-set 1 was a medium of constant velocity, data-set 2 had medium velocity varying spatially in the vertical direction (so a pseudo-stratified velocity), and data-set 3 is from the famous Marmousi model\footnote{The Marmousi model data-set was created in 1988, and has become an industry standard data-set for testing seismic imaging techniques. Originally created by the Institut Francais du P´etrole, the data is synthetic, and is based on the sub-terrain structure found in the North Quenguela trough in the
Cuanza basin}, which has velocities varying in all spatial directions.

In order to calculate sufficient ICs for training, finite difference methods were used for the first $T_1$ training steps to create a discrete wavefield. Therefore, the boundary data is from the set $x_i\in[0,X_{\textrm{max}}]$ and $t_i\in[0,T_1]$, where $T_1 \ll T_{\textrm{max}}$. For the constant and stratified data-sets, $T_1 = 0.02$ seconds, which was the first 10 time steps. For the Marmousi data-set, $T_1 = 0.04$ seconds, so the first 20 times steps were used, as were multiple simulations from multiple sources.

The training was split into two distinct phases: the boundary phase and the physics phase. Phase 1 allows the network to prioritise minimising the boundary loss term without the interference of the physics loss. Half way through the iterations (at step 500'000) the physics loss term was introduced, allowing the network to minimise over the rest of the domain.

\subsubsection{Result}

A full FD simulation was run for each data-set, which represents the `ground truth', and the error is the difference between the FD model and the PINN model. The aim of the study was to see how well a PINN could model wave propagation through a known 2D velocity domain, and how much better it performed vs a standard ANN.

Standard ANN have tended to capture the outward propagating wave, but usually fail to capture more subtle behaviours such as sheer, reflection, and refraction, and thus cannot generalise the solution far outisde the training domain (so for $t > T_{\mathrm{max}}$). This problem is significantly lessened by introducing the physics loss.

\begin{figure}[t]
\begin{center}
\includegraphics[width=0.8\textwidth]{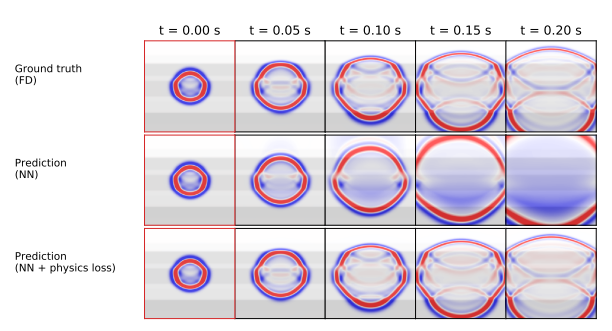}
\caption{Wave behaviour estimate for the stratified data-set using 3 methods: FD, ANN, PINN. The ANN fails to capture many of the reflected wave behaviours at larger $t$, but does model the leading behaviour}
\label{di:wavebeh}
\end{center}
\end{figure}

 The errors in the ANN approach tend to occur at fault lines (where the velocity changes significantly). This is likely because the ANN is given no information for how the wave field should change through time and space as the velocity of the medium changes, and so cannot learn the properties of reflected waves, etc, without sufficient data (and this data is often lacking in real applications).

\begin{figure}[b]
\begin{center}
\includegraphics[width=0.8\textwidth]{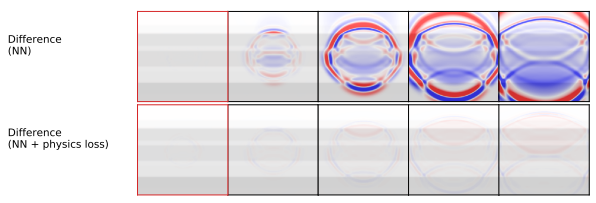}
\caption{The difference between the FD simulation and each of the two networks. The error between the PINN and the FD simulation is almost negligible}
\label{di:waveerr}
\end{center}
\end{figure}

 The authors highlight that the greatest advantage of using a PINN over a FD simulation is that the solution is much better generalised. As an example, for the 3rd data-set multiple sources were used to create a PINN. If one wanted to estimate the wavefield generated from a new source for the same domain at, say, $t=0.4$, the FD method would require simulating all time over all space up to the required time using the new source term. Due to the fact that the PINN solution is more generalised, further simulations are not required to find this value, and one can simply entire the time, space, and source information as required. This means that using a trained PINN is in the order of $\times1000$ faster than using FD.

The PINN approach is also easily scalable, as adding dimensions, changing data-sets, and altering the physics is very simple to do. However, the discontinuities in the medium velocity can still cause errors to appear in the solution, though this is also a problem for FD methods. To avoid this issue, the 3rd data-set was smoothed slightly where two different velocities interface. There are three potential solutions to this problem:
\begin{enumerate}
    \item Use individual PINNs for each section of the subsurface terrain, which would allow for discontinuities in the medium.
    \item Introduce a loss term that allows for wavefield seperation.
    \item Use a more intelligent sampling scheme for the data, so areas of high velocity change are sampled more frequently during minimisation, as dicussed in the previous chapter.
\end{enumerate}
 The conclusion, however, is that PINNs can learn many of the intricate behaviours that waves exhibit, can create a continuous solution that can accurately estimate values well outside of the training data, and are extremely quick to use once trained. 

\subsection{The Inverse Problem}

Since it has been shown that PINNs can find accurate solutions for the wavefield, the more crucial question is whether or not they can be used as a substitute to solve the inverse problem - finding the distribution of medium velocity. Chao Song et al. from King Abdullah University (2021) \cite{inversewave} applied the PINN described in chapter 3 to better inform the WRI method.

\subsubsection{Method}

The key idea behind this study is that two PINNs were used to reconstruct the velocity field.

\begin{enumerate}
    \item The first PINN reconstructed the scattered wavefield on a per frequency basis.
    \item The second PINN used the reconstructed wavefield from the first PINN to make changes to the velocity field (squared slowness).
\end{enumerate}

 Instead of using the time-domain wave equation, the study utilised the Lippmann Schwinger form of the acoustic wave equation \cite{scatwave} 
\begin{equation}
    \omega^2m\delta u + \nabla^2 \delta u = i\omega^2 \delta m u_0
\end{equation}
which finds the scattered wavefield $\delta u = u - u_0$ for differing angular frequencies $\omega$, where $u$ is the wavefield and
\begin{equation}
    u_0(x) = iH_0^{(1)}(\omega \sqrt{m_{n0}}|x-x_s|)
\end{equation}
for isotropic, 2D domains, is the background wavefield, with $x_s$ being a point source location. $\delta m = m-m_0$ is the perturbed squared slowness, with $m_0$ being the slowness of the homogeneous earth model. $H_n^{(k)}$ is a kth kind Hankel function, defined as
\begin{equation}
    H_0^{(1)}(z) = \frac{1}{i \pi}\int_0^\infty \frac{e^{(\frac{z}{2})(\frac{t-1}{t})}}{t}dt
\end{equation}
which has real and imaginary parts. 

Using this set of equations, the physics-informed loss function (which is referred to as MSE in this particular piece of literature) for the forward problem was defined in the standard way of a balance between the physics and the data
\begin{equation}
\begin{aligned}
    \textrm{MSE}_f = \sum_{is}^{N_s}\bigg(\frac{1}{N_r}\sum_{i=1}^{N_r}||\delta d^i - \delta u(x_r^i)||_2^2 + \\
    \frac{\alpha}{N}\sum_{i=1}^N||\omega^2 m_1^i \delta u^i + \nabla^2 \delta u^i + \omega^2(m_1^i - m_0^i)u_0^i||^2_2\bigg)
    \end{aligned}
\end{equation}
where the first term is the data loss term, and the second term encapsulates the physics of the system (via the Lippmann Schwinger equation). The adjustable parameters in the equation define the behaviour of the scattered wavefield $\delta u$. Each $is$ is a source location, $\delta d = d^i - u_0(x=x_r)$ is recorded scattered data, and $m_1$ is the initial squared slowness model. $\alpha$ is a weight term, which was set to 0.00001 (simply through trial and error testing), and $N$ is the number of points selected for the physics constraint.

The physics informed loss function for the inverse problem was defined as
\begin{equation}
    \begin{aligned}
        MSE_b = \frac{1}{N_p}\sum_{i=1}^{N_p}||\omega^2 m^i \delta u^i + \nabla^2 \delta u^i + \omega^2(m^i - m_0^i)u_0^i||^2_2 + \\
        \epsilon\sqrt{\bigg(\frac{\partial m^i}{\partial x}\bigg)^2+\bigg(\frac{\partial m^i}{\partial z}\bigg)^2}
    \end{aligned}
    \label{eq:mseb}
\end{equation}
which, noticeably, has no data term. The second term, called the total-variation term (TV), was added to stabilise the training process, with $\epsilon=0.1$. Here, the scattered wave $\delta u$ is the PINN solution for a fixed frequency. The inverse PINN uses the forward PINN to then make changes to the earth model $m$. In truth, the algorithm is deceptively simple.

\begin{algorithm}[t]
\caption{PINN-based WRI}\label{alg:PINNWRI}
\begin{algorithmic}
\Require $d$ \Comment{The observed data-set}
\Require $m_0$ \Comment{The background squared slowness model}
\Require $m_1$ \Comment{An initial squared slowness model}
\Require $u_0$ \Comment{The background wavefield}
\Require $n$ \Comment{Maximum number of iterations}
\Require $[f_{\textrm{min}} : df : f_{\textrm{max}}]$ \Comment{Selected angular frequencies}
\Require $\alpha$ \Comment{Weight for physics term in forward solver}
\For{$f = [f_{\textrm{min}} : df : f_{\textrm{max}}]$}
    \For{$i=1:n$}
        \State Reconstruct the scattered wavefield for fixed frequency
        \State Reconstruct velocity model using reconstructed scattered wavefield
        \EndFor
\EndFor
\State \Return Inverted velocity model
\end{algorithmic}
\end{algorithm}

The velocity model is therefore continually improved by modelling the scattered wavefield of differing frequencies. 

Due to the complexity of the task, size of the domain, and dynamic nature of the domain, the networks used in this study were quite large. The forward (scattered wavefield) PINN used a descending hierarchy structure with 8 hidden layers, ordered as $[128, 128, 64, 64, 32, 32, 16, 16, 8, 8]$, whereas the inverse (velocity) PINN used a flat structure with 8 hidden layers, ordered as $[20, 20, 20, 20, 20, 20, 20, 20]$. The input for the scattered wavefield PINN was $\mathbb{R}^3$, containing a 2D location $x$ and a source term $x_s$, and the output was $\mathbb{R}^2$ which were the real and imaginary parts of the scattered wavefield. The input for the velocity model PINN was a 2D point, and the output was simply the velocity.

The networks were trained using the 20'000 iterations of the ADAM optimisation algorithm, followed by 150'000 iterations of L-BFGS.

\subsubsection{Data}

The method was tested on two data-sets, but we will focus on 2D Marmousi model for two reasons
\begin{enumerate}
    \item The data-set was used in the previous study, so gives a good point of comparison.
    \item It was the more complex of the two data-sets, and so is more interesting to analyse.
\end{enumerate}

 The data consisted of
\begin{enumerate}
    \item 10 sources, spaced evenly on the surface of the terrain.
    \item 301 receivers, spaced evenly at a depth of 25m, which collect data for the data loss term.
    \item 80'000 randomly sampled points inside the domain, which collect points for the physics loss
\end{enumerate}

 The initial velocity model was generated in such a way that the velocity increased as z increase (so a stratified model, not too dissimilar to the second data-set used in the previous study).

\begin{figure}[t]
\begin{center}
\includegraphics[width=0.9\textwidth]{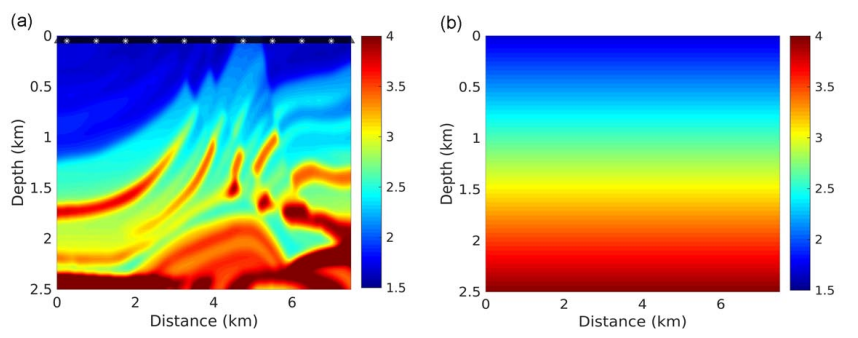}
\caption{Left: True Marmousi velocity model. Right: Initial velocity model for PINN WRI algorithm}
\label{di:mar_init}
\end{center}
\end{figure}

\subsubsection{Result}

The resulting velocity field after performing PINN-based WRI for 3Hz, 4Hz, and 5Hz (3 sets of angular frequencies) produced an earth model that had some characteristics of the true solution, but was far too smooth. This follows from the experimental data that was analysed in the previous chapter. The error here is caused by the scattered wavefield being too smooth. In other words, areas with large changes in derivative had a tendency to have high errors. However, despite the fact that the velocity model from PINN-based WRI wasn't a fantastic approximation of the true solution, it did prove to be incredibly useful. 

In section 5.1.1, the problems and shortcomings of FWI and WRI were discussed in detail. One of these shortcomings is that the the minimisation problem is often plagued by local minimums, and that the final solution is dependent on the initial earth model provided. Producing a good inital earth model is incredibly challenging, and is often somewhat impossible. As such, a stratified initial mode (such as in figure \ref{di:mar_init}) is often used. 

 The solution provided by the PINN-based WRI, whilst being too smooth to realise high resolution details, does capture some of the general behaviour of the true 2D Marmousi model. It is certainly a better fit than the initial velocity model. The final velocity model created by the PINN-based WRI was then used as the initial earth model for FWI, and its solution was compared to the solution of using the stratified data as the initial model.

\begin{figure}%[t]
\begin{center}
\includegraphics[width=0.9\textwidth]{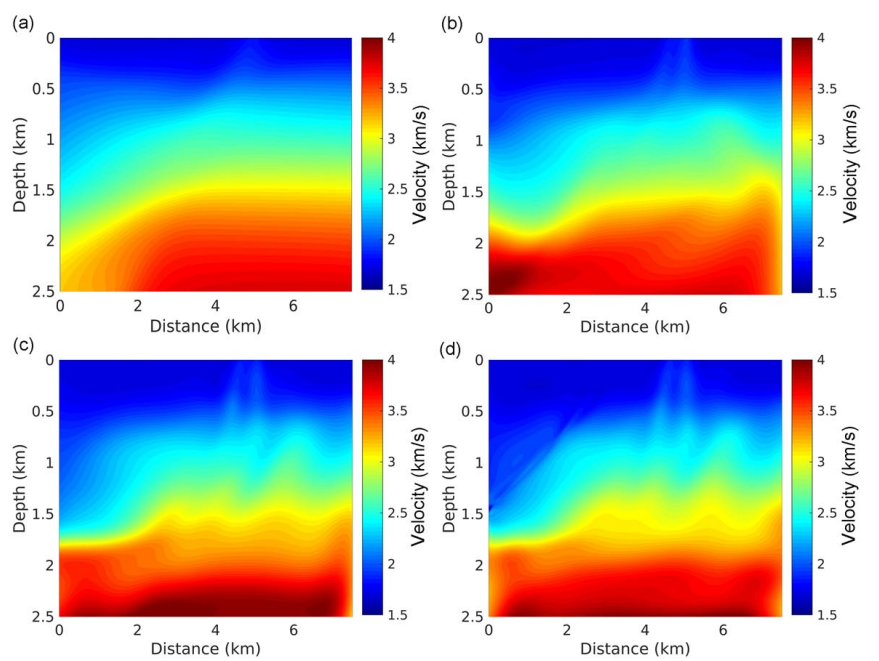}
\caption{Top left: PINN velocity model after 1 iteration using 3Hz. Top right: PINN velocity model after 4 iterations of 3Hz. Bottom left: PINN velocity model after 4 iterations of 4Hz. Bottom right: Pinn velocity model after 3 iterations of 5Hz}
\label{di:pinn_vel1}
\end{center}
\end{figure}

\begin{figure}%[t]
\begin{center}
\includegraphics[width=0.9\textwidth]{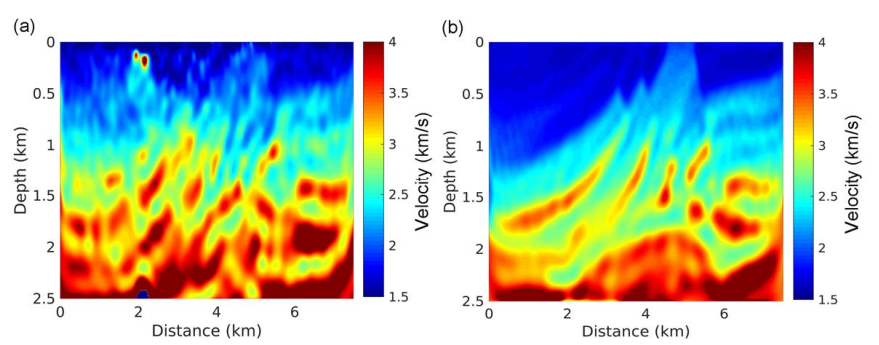}
\caption{Left: Estimated velocity model from stratified initial velocity model using FWI. Right: Estimated velocity model from PINN solution initial velocity model using FWI.}
\label{di:pinn_vel2}
\end{center}
\end{figure}

 Using the PINN WRI solution as the initial model gave a significantly better final velocity model than using the stratified model. It still appears a little too smooth, but the main features of the solution certainly match the true data.

In comparing the PINN method with other numerical techniques, it was commented that, whilst FD methods work quickly for small scale problems, they are not well suited to large scale, high dimensional problems. This is because they cannot be calculated in parallel like a PINN algorithm can be. Further more, adding more complexity to the properties of the domain, such as having vertical transversely isotropic media, or isotropic elastic media dramatically increases the size of the impedance matrix used for FD methods.

Finally, as was mentioned in the previous study, the flexibility of this method cannot be understated. In order to make WRI and FD methods work for a higher dimensional domain, where the sub-surface structure is more complex (eg, has fluid-saturated porous sections), the algorithms would need huge modifications, whereas the PINN approach simply requires altering the loss function to include differing equations.

The major shortcomings here are that the scattered wavefield solutions are overly smooth, which in turn leads to an inverted velocity model that is also too smooth. One solution is to potentially increase the size of the network, allowing for more degrees of freedom, or use an individual network for each frequency (which could then be done in parallel). However, this also obviously increases computational cost considerably.

\chapter{Conclusion}
Due to the creation of packages like Tensorflow and Pytorch in python, and flux in Julia, creating artificial neural networks has never been easier. On top of this, memory has become incredibly inexpensive and, because of the parallel nature of many of the operations when using a neural network, large and powerful GPUs can be easily leveraged to achieve incredibly fast computing times. As such, minimising the loss function for high dimensional problems with many parameters is significantly faster than it was just a few years ago.

Thanks to advances in computational differentiation, it has been shown that the loss function for artificial neural networks can be created in such a way as to accurately approximate the solution for physical systems in parts of the domain where no data exists. Further more, these \textit{universal approximators} can be cleverly designed in order to have explicit access to known behaviours, such as ICs and BCs.

The solutions that physics-informed neural networks produce are continuous functions. They therefore have the superior characteristic of being able to produce values for inputs that are outside of the domain or beyond known data, unlike other numerical methods, and they do not require any further calculations in order to refine the solution (such as smaller steps sizes in finite difference methods, or a finer mesh in finite element analysis).

Furthermore, a deep dive into error analysis has shown how, why and where errors may occur in a physics-informed neural network approximation. These errors tend to occur for two reasons:
\begin{enumerate}
    \item The nearest pocket of dense data is relatively far away.
    \item The relative derivatives in this part of the domain are large.
\end{enumerate}

 These errors occurred both in the ideal wave system in chapter 4, and the wavefield reconstruction inversion method in chapter 5. As such, we can estimate points of large error in the approximate solution, and potentially employ data sampling techniques to improve the accuracy of these areas using the physics loss term. This pattern of high derivative and high error is consistent across recent studies, both in the forward problem and in the inverse problem, and is the main limitation when using physics-informed neural networks

It is not suggested that physics-informed neural networks replace conventional methods for solving PDEs, but instead act as another tool one could use to solve a problem or even validate a solution. As is shown in chapter 5, PINN solutions certainly have there place in solving real world problems, and these uses still require deeper and broader experimentation to find the limit of their applications. These networks are particularly useful in complex systems as they are incredibly flexible and easy to modify to incorporate new physics, new data, or new dimensions.

Further work certainly needs to be done in error analysis and convergence rate. The groundwork is laid out in this thesis, but still requires more rigorous proof and theory. A better understanding of how a solution is converged to could prevent problems such as local minimums, which already plague many of the optimisation problems we face today. Errors clearly manifest themselves in pre-defined ways, and as such they could be tackled more effectively during the minimisation process, rather than ex post facto.

% Comment the following THREE lines if you do NOT have an Appendix
\appendix
\chapter{}

\section{Proofs and Examples}

\subsection{Expression Swell}
ANNs are, fundamentally, activation functions $\sigma: x \to [-1,1]$
that are continually summed and embedded into each other, sometimes thousands of times. Expression swell is a phenomena in computing where by, as the calculation progresses, the size of the problem grows exponentially.

Take the following, simple function
\begin{equation}
    f(x) = \tanh\left(\sin\left(\cos\left(x\right)\right)\right)
\end{equation}
which has 3 operations. Clearly, if it was an ANN, it would be incredibly simple. Taking the derivative with respect to $x$ yields
\begin{equation}
    f'(x) = -\sin\left(x\right)\cos\left(\cos\left(x\right)\right)\operatorname{sech}^2\left(\sin\left(\cos\left(x\right)\right)\right)
\end{equation}
which has 9 operations. We again differentiate to get
\begin{equation}
    \begin{aligned}
        f''(x) &= -2\sin^2\left(x\right)\cos^2\left(\cos\left(x\right)\right)\operatorname{sech}^2\left(\sin\left(\cos\left(x\right)\right)\right)\tanh\left(\sin\left(\cos\left(x\right)\right)\right)\\&-\sin^2\left(x\right)\sin\left(\cos\left(x\right)\right)\operatorname{sech}^2\left(\sin\left(\cos\left(x\right)\right)\right)\\&-\cos\left(x\right)\cos\left(\cos\left(x\right)\right)\operatorname{sech}^2\left(\sin\left(\cos\left(x\right)\right)\right)
    \end{aligned}
\end{equation}
which has 27 operations. We again differentiate to get

\begin{equation}
    \begin{aligned}
        f'''(x) &= -\operatorname{sech}^2\left(\sin\left(\cos\left(x\right)\right)\right)\left(4\sin^3\left(x\right)\cos\left(\cos\left(x\right)\right)\right)\sin\left(\cos\left(x\right)\right)\tanh\left(\sin\left(\cos\left(x\right)\right)\right) \\
        &+ 4\cos\left(x\right)\sin\left(x\right)\cos^2\left(\cos\left(x\right)\right)\tanh\left(\sin\left(\cos\left(x\right)\right)\right)\\
        &-2\sin^3\left(x\right)\cos^3\left(\cos\left(x\right)\right)\operatorname{sech}^2\left(\sin\left(\cos\left(x\right)\right)\right)\\
        &+3\cos\left(x\right)\sin\left(x\right)\sin\left(\cos\left(x\right)\right)-\sin^3\left(x\right)\cos\left(\cos\left(x\right)\right)-\sin\left(x\right)\cos\left(\cos\left(x\right)\right)\\
        &-2\sin\left(x\right)\cos\left(\cos\left(x\right)\right)\operatorname{sech}^2\left(\sin\left(\cos\left(x\right)\right)\right)\tanh\left(\sin\left(\cos\left(x\right)\right)\right)\\
        &\left(2\sin^2\left(x\right)\cos^2\left(\cos\left(x\right)\right)\tanh\left(\sin\left(\cos\left(x\right)\right)\right)\right)+\sin^2\left(x\right)\sin\left(\cos\left(x\right)\right)\\
        &+\cos\left(x\right)\cos\left(\cos\left(x\right)\right)
    \end{aligned}
\end{equation}
which has... a lot more than 3 operations! 27 operations for a computer is not very many, but large ANNs can have thousands of fully connected nodes, and so $f(x)$ could have millions operations alone. This size increases dramatically during symbolic differentiation.

\subsection{Solution to 1D Wave Equation}

1D waves of all kinds with a multitude of initial and BCs have been studied for centuries, and they therefore make for a nice problem to test PINNs because solutions are well understood. For 1D waves with Dirichlet BCs, the solution can be found by D'Almbert's method, Fourier transform, or separation of variables. Here we will show that the solution to the problem described in (insert ref here) is the series solution (insert ref here) using separation of variables.

The 1D wave equation takes the following form
\begin{equation}
    \frac{\partial^2 u }{\partial t^2} = c^2 \frac{\partial^2 u}{\partial x^2}
\end{equation}
We search for a solution that takes the form $u(x, t) = X(x)T(t)$, which yields
\begin{equation}
    \frac{1}{X}\frac{d^2 X}{dx^2} = \frac{1}{c^2}\frac{1}{T}\frac{d^2}{dt^2}
    \label{eq:waveeq}
\end{equation}
by substituting in the expected form into \ref{eq:waveeq}. Since both sides are functions of a different variable, they must be constant, and so
\begin{equation}
    \begin{aligned}
        \frac{1}{X}\frac{d^2 X}{dx^2} &= -k^2 \\
        \frac{1}{c^2}\frac{1}{T}\frac{d^2}{dt^2} &= -k^2
    \end{aligned}
\end{equation}
These can then be solved as coupled ODEs. Solving for $X(x)$ and $T(t)$ gives
\begin{equation}
    \begin{aligned}
        X(x) &= A\cos(kx) + B\sin(kx) \\
        T(t) &= C\cos(kct) + D\cos(kct)
    \end{aligned}
\end{equation}
Applying the BCs finds the constants in the solutions. If
\begin{equation}
    u(x, t) = (A\cos(kx) + B\sin(kx))(C\cos(kct) + D\cos(kct))
\end{equation}
then using $u(0, t) = 0$ gives $A = 0$ and $k=\frac{n\pi}{2}$, where $n$ is a positive integer. This gives a particular solution
\begin{equation}
    u_n(x, t) = \big(\alpha_n\cos(k_nct) + \beta_n\cos(k_nct)\big)\sin\bigg(\frac{n\pi x}{2}\bigg)
\end{equation}
The IC gives $u_t(x, 0) = 0$, and so $\beta_n = 0$, giving
\begin{equation}
    u_n(x, t) = \alpha_n\cos(k_nct)\sin\bigg(\frac{n\pi x}{2}\bigg)
\end{equation}
The orthogonality of the sine function means that, for a string of length $2$, we have
\begin{equation}
\begin{aligned}
    \alpha_n &= \int_0^2 u(x, 0)\sin\bigg(\frac{n\pi x}{2}\bigg)dx \\
    &=\int_0^2 x(2-x)\sin\bigg(\frac{n\pi x}{2}\bigg)dx \\
    &=-\dfrac{8{\pi}n\sin\left({\pi}n\right)+16\cos\left({\pi}n\right)-16}{{\pi}^3n^3}
    \end{aligned}
\end{equation}
with $ck_n = \frac{cn\pi}{2} = \frac{n\pi}{2}$ (for $c=1$), which gives the series solution

\begin{equation}
    u(x, t) = \sum_{n=1}^\infty -\frac{8{\pi}n\sin\left({\pi}n\right)+16\cos\left({\pi}n\right)-16}{{\pi}^3n^3} cos\bigg(\frac{n\pi t}{2}\bigg)\sin\bigg(\frac{n\pi x}{2}\bigg)
\end{equation}

\section{Code for Errors and Plotting}

All code for plots and figures is kept in:

\begin{center}
   \url{https://github.com/Teddyzander/PhysicsInformedNeuralNetwork}
\end{center}

% If you need more than one appendix, then just use another \chapter command
%\chapter{Yet Another Appendix}

\newpage

\end{document}